\documentclass[letterpaper]{article} 
\usepackage[preprint]{aaai2027}
\usepackage[hyphens]{url}  
\usepackage{graphicx} 
\usepackage{caption} 
\usepackage{algorithm}
\usepackage{algorithmic}

\usepackage{newfloat}
\usepackage{listings}
\DeclareCaptionStyle{ruled}{labelfont=normalfont,labelsep=colon,strut=off} 
\floatstyle{ruled}
\newfloat{listing}{tb}{lst}{}
\floatname{listing}{Listing}

\usepackage{booktabs}
\usepackage{amsmath}
\usepackage{amssymb}

\usepackage{multirow}
\providecommand{\equalcontrib}{\textsuperscript{*}}

\title{Correcting What You Cannot See: Credit Assignment for Perception Distillation in Multimodal Reasoners}
\author{
Feng Xiong\equalcontrib,
Leyan Xue\equalcontrib,
Hongyu Lin\\
\normalfont\small\textsuperscript{*}Equal contribution
}

\begin{document}

\maketitle

\begin{abstract}
On-policy distillation provides dense supervision for multimodal reasoners, but its trajectory-level reward cannot determine whether a failed answer arose from perception or subsequent reasoning. Perception Success Rate (PSR), estimated from multiple reasonings sharing one perception, remains ambiguous because low success conflates perceptual insufficiency with reasoning difficulty. We introduce \textbf{Perception-Correction Distillation (PCD)}, a label-free method that identifies correctable perception failures using downstream failure and teacher--student disagreement as complementary witnesses. Their product, $(1-\mathrm{PSR})\widetilde{\mathrm{KL}}$, forms a soft AND gate that strengthens distillation only when both witnesses are present. We motivate this rule through Bayesian evidence combination and show that multiplication is the unique normalized bilinear gate that vanishes when either witness is absent. PCD uses separated perception--reasoning rollouts and mean-preserving weights, leaving the reasoning objective unchanged. Across eight benchmarks, PCD improves the 8B$\rightarrow$2B macro average from 44.50 with OPD to 47.28 and the 32B$\rightarrow$8B result from 56.94 to 61.22. In matched 2B ablations, removing PCD and separated rollout reduces held-out average by 2.22 and 0.88 points, respectively. Effective multimodal distillation therefore depends not only on what the teacher predicts, but also on identifying when perception is the appropriate target of correction.
\end{abstract}

\section{Introduction}
\label{sec:introduction}

Multimodal large language models (MLLMs) must first extract evidence from an
image and then reason over it. Deployable students often acquire these
capabilities through knowledge distillation~\cite{hinton2015distilling,
furlanello2018born,beyer2022patient}: a larger teacher supplies richer targets
than task labels alone. For generated language, sequence-level and rationale
distillation transfer complete outputs or intermediate explanations
~\cite{kim2016sequence,hsieh2023distilling}. On-policy distillation (OPD)
instead evaluates trajectories sampled by the current student, aligning
teacher supervision with the states the student actually visits
~\cite{gu2024minillm,agarwal2024,lu2025onpolicy}. This property is attractive
for visual reasoning because the student's own visual mistakes become
training examples.

Yet a multimodal response is not a homogeneous token sequence. A short span
may record the decisive observation---a marked angle, spatial relation, or
chart value---while a much longer span derives the answer. Visual
chain-of-thought and multimodal-rationale methods increasingly expose this
separation~\cite{chen2024vctp,shao2024visualcot,he2024multimodal,
wang2024tsciq,cheng2025comt}. It reveals two known mismatches. First,
\emph{token dilution} lets long derivations dominate the few image-dependent
tokens; VPPO addresses this with token-level visual focusing
~\cite{huang2026vppo}. Second, \emph{objective mismatch} suggests imitating a
strong visual teacher for perception while allowing reasoning to explore
under verifiable reward~\cite{shao2024grpo,yu2025dapo,deepseekai2025r1}. We
therefore use an \texttt{<aware>} perception span trained by distillation and
a \texttt{<cot>} reasoning span trained by RL.

\begin{figure*}[t]
    \centering
    \includegraphics[width=\textwidth,pagebox=cropbox]{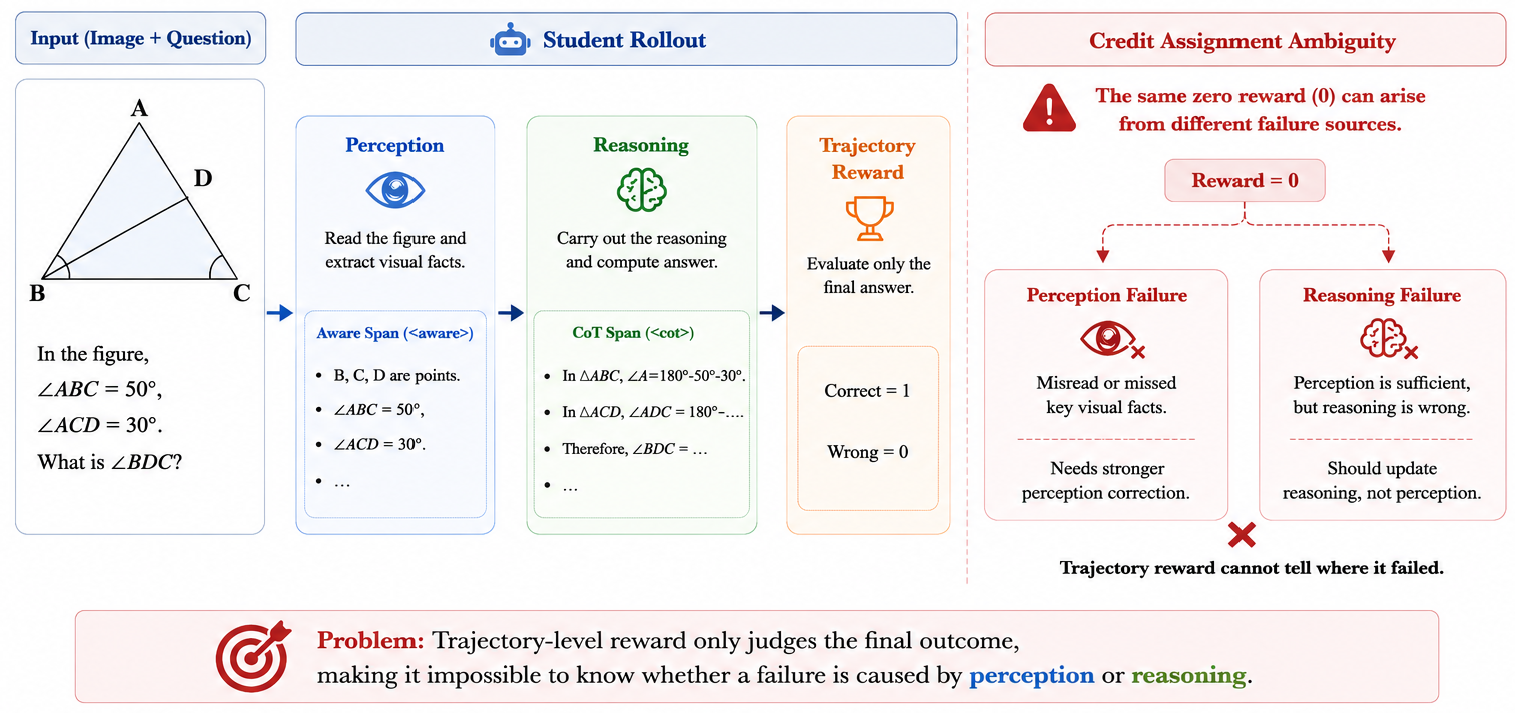}
    \caption{Trajectory reward maps perception and reasoning failures to the
    same outcome. PCD adds teacher disagreement as a second witness on the
    perception span.}
    \label{fig:credit_ambiguity}
\end{figure*}

This decomposition exposes a third, unresolved mismatch:
\emph{trajectory-level credit ambiguity}. A verifier scores only the completed
response. The same zero reward can follow either a misread image or an
incorrect derivation from an adequate observation (Figure~\ref{fig:credit_ambiguity}).
These failures require different updates: the former warrants stronger
perception correction, whereas the latter should primarily change reasoning.
Outcome-based RL cannot distinguish them, and uniform OPD corrects every
perception regardless of whether perception caused the failure.

Existing credit signals do not recover the missing attribution. PPO, GRPO,
and DAPO assign outcome-based advantages to trajectories rather than causal
stages~\cite{schulman2017ppo,shao2024grpo,yu2025dapo}. Process reward models
localize errors within derivations but require step supervision or an
additional verifier, and normally judge reasoning rather than its visual
premise~\cite{lightman2024verify,wang2024mathshepherd,wang2025visualprm}.
Delayed-return decomposition can move reward toward earlier actions
~\cite{arjona2019rudder}, but cannot reconstruct information absent from the
outcome: whether the initial observation was sufficient.

A natural baseline samples several reasonings from one perception and uses
their mean reward as a \emph{Perception Success Rate} (PSR). PSR estimates
downstream value under the current reasoner, but not perception quality. Low
PSR may indicate a bad perception, a difficult problem, or weak reasoning.
Additional samples reduce variance around this confounded quantity without
identifying its cause. Consequently, success-only weighting may suppress
teacher supervision for a correct perception merely because its derivations
failed.

We ask instead: \emph{which failed perceptions are both plausibly deficient
and teacher-correctable?} We combine downstream failure, $1-\mathrm{PSR}$,
with student--teacher disagreement on the perception span,
$\widetilde{\mathrm{KL}}$:
\begin{equation}
    d_i=(1-\mathrm{PSR}_i)\widetilde{\mathrm{KL}}_i.
\end{equation}
This product is a conservative soft AND gate. Neither failure alone nor
disagreement alone triggers strong correction; only their conjunction does.
We derive this interaction from conditionally independent likelihood-ratio
witnesses and show that multiplication is the unique normalized bilinear gate
that vanishes when either witness is absent.

The resulting method, \textbf{Perception-Correction Distillation (PCD)}, first
samples $a$ perceptions and then $b$ reasoning continuations from each fixed
perception. This separated rollout makes perception the unit of estimation.
PCD applies a mean-preserving weight only to perception distillation, thereby
reallocating a fixed teacher-supervision budget while leaving the reasoning
objective and group-relative advantages unchanged. It requires no perception
labels, learned gate, or additional model beyond the teacher already used by
OPD.

Our contributions are:
\begin{itemize}
    \item We formulate perception distillation as an identifiability problem
    and prove that reward-only PSR cannot separate perception sufficiency from
    reasoning difficulty.
    \item We derive a two-witness multiplicative deficiency score and show
    that its mean-normalized weighting follows the optimal first-order
    reallocation direction under a fixed supervision budget.
    \item We introduce separated perception--reasoning rollout, distilling
    perception while optimizing reasoning with verifiable-reward RL.
    \item Across eight benchmarks, PCD reaches 47.28 for 8B$\rightarrow$2B
    and 61.22 for 32B$\rightarrow$8B transfer; matched ablations lose 2.22
    points without PCD weighting and 0.88 without separated rollout.
\end{itemize}
\section{Related Work}

\paragraph{Knowledge and on-policy distillation.}
Knowledge distillation transfers a teacher's predictive distribution to a
smaller student~\cite{hinton2015distilling,beyer2022patient}; its benefits can
also persist without a capacity reduction~\cite{furlanello2018born}.
Sequence- and rationale-level methods train on teacher-generated outputs
~\cite{kim2016sequence,hsieh2023distilling}, whereas recent language-model
methods reduce distribution mismatch by learning from student-generated
trajectories. MiniLLM minimizes reverse KL on student samples
~\cite{gu2024minillm}; generalized on-policy distillation studies alternative
divergences and student sampling~\cite{agarwal2024,lu2025onpolicy}. Policy
distillation similarly transfers action distributions in RL
~\cite{rusu2016policy,czarnecki2019distilling}. PCD retains the on-policy
learner but changes how perception-span supervision is allocated across
trajectories, rather than aligning heads, relational knowledge, or unequal
visual-token spaces~\cite{zhao2024multihead,yang2025multimodalkd,feng2026emkd}.

\paragraph{Multimodal perception and visual reasoning.}
Recent technical reports emphasize native-resolution perception and
test-time reasoning in open MLLMs~\cite{wang2024qwen2vl,bai2025qwen25vl,
chen2024internvl25,bai2025qwen3vl}.
Visual CoT explicitly localizes relevant image regions before
reasoning~\cite{shao2024visualcot}, illustrating that perception and
reasoning can benefit from distinct intermediate representations. Related
AAAI work aligns multimodal evidence with language thoughts in a latent
reasoning space~\cite{he2024multimodal}, distills teacher-generated multimodal
rationales into smaller models~\cite{wang2024tsciq}, and uses visual
chain-of-thought prompting to select evidence for knowledge-based reasoning
~\cite{chen2024vctp}; CoMT and KAM-CoT study longer or knowledge-grounded
multimodal thought~\cite{cheng2025comt,mondal2024kamcot}.
Vision-OPD~\cite{yuan2026visionopd} directly targets the regional-to-global
perception gap through on-policy self-distillation: the same MLLM, conditioned
on a relevant image crop, acts as a privileged teacher for a full-image
student and supplies token-level distribution supervision on the student's
rollouts. Vision-OPD transfers fine-grained regional evidence into the
full-image policy, whereas PCD addresses a complementary credit-assignment
problem: it combines downstream failure with teacher--student disagreement to
decide which sampled perception trajectories should receive stronger
distillation.
VPPO~\cite{huang2026vppo} measures token-level visual
dependence and introduces
Token Gradient Filtering and Trajectory Advantage Shaping. PCD is
complementary: VPPO asks which \emph{tokens} are visually grounded, while PCD
asks which \emph{perception trajectories} are plausible teacher-correctable
failures. Their weights act at different granularities and can be multiplied.

\paragraph{Reinforcement learning for reasoning.}
PPO~\cite{schulman2017ppo} provides the clipped policy-optimization foundation
used by many language-model RL systems. GRPO removes the learned critic through
group-relative normalization~\cite{shao2024grpo}, while DAPO adds stability
and token-level optimization refinements~\cite{yu2025dapo}; DeepSeek-R1
demonstrates the broader capability gains obtainable from verifiable-reward
RL~\cite{deepseekai2025r1}. These
methods optimize complete reasoning trajectories from outcome rewards. PCD
does not replace their reasoning objective; it uses the same outcomes to
decide where teacher supervision on the preceding perception is most useful.

\paragraph{Process supervision and credit assignment.}
Outcome-only feedback cannot directly identify erroneous intermediate steps.
Process reward models address this limitation with step-level
supervision~\cite{lightman2024verify}, and
Math-Shepherd constructs such supervision automatically for mathematical
reasoning~\cite{wang2024mathshepherd}. VisualPRM extends this approach to
multimodal process evaluation~\cite{wang2025visualprm,zhou2026visualprm}.
In general RL, RUDDER, STAS, and Latent Reward redistribute delayed feedback
~\cite{arjona2019rudder,chen2024stas,qu2025latent}. PCD differs in target: it assigns
credit from the final reasoning outcome back to a latent \emph{perception}
choice, then uses that credit to reweight distillation rather than to create a
new reward. Because outcome reward alone cannot identify the failure source,
PCD adds teacher disagreement as an independent witness.

\paragraph{Perception-aware weighting.}
Within a separated perception--reasoning rollout, a natural baseline is the
Perception Success Rate (PSR): generate multiple reasonings from one
perception and weight distillation by their mean success. PSR treats
low-success perceptions as unreliable. PCD instead distinguishes two cases:
teacher-aligned low-success perceptions receive no additional correction,
whereas low-success perceptions that also disagree with the teacher are
up-weighted. Thus PSR and PCD encode contrasting interpretations of
unsuccessful trajectories and provide a direct test of whether teacher
disagreement resolves reward-only credit ambiguity.
\section{Credit Assignment and Perception-Correction Distillation}
\label{sec:method}

\begin{figure*}[t]
    \centering
    \includegraphics[width=\textwidth]{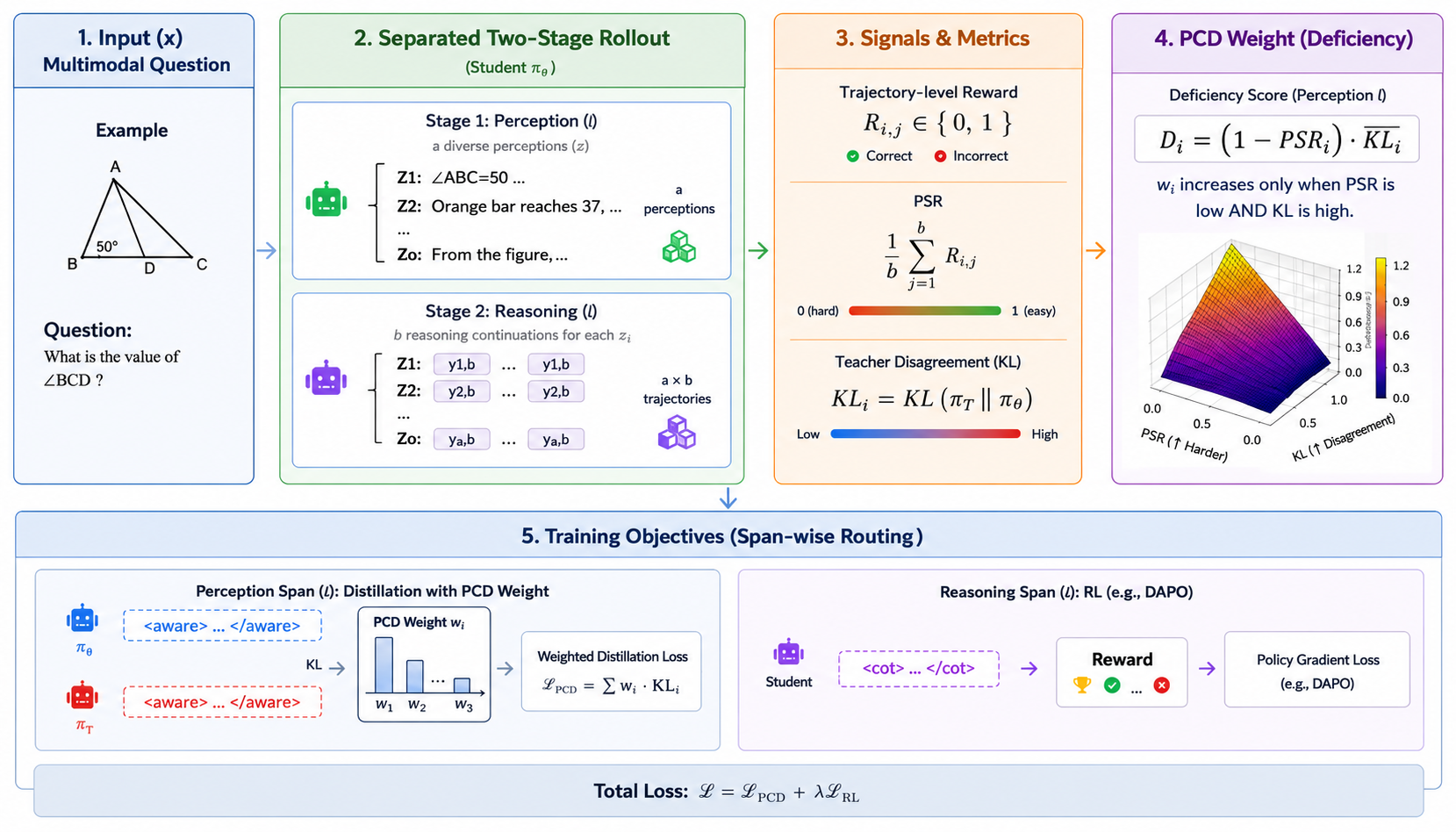}
    \caption{Overview of the separated rollout, divided objectives, and PCD weighting.}
    \label{fig:framework}
\end{figure*}

We first formalize the perception-credit problem and then present its implementation in PCD. The central latent variable is whether a sampled perception contains enough visual information for the current reasoner to solve the problem.

\subsection{Problem Formulation}

Given an image--question pair $x$, we factor the student policy into perception and reasoning stages,
\begin{equation}
    \pi_\theta(y,z\mid x)=\pi_p(z\mid x)\pi_r(y\mid x,z),
    \label{eq:factorization}
\end{equation}
where $z$ is enclosed by \texttt{<aware>} tags and $y$ by \texttt{<cot>} tags. Let $R(x,z,y)\in[0,1]$ denote the verifiable reward of the completed response. The frozen teacher $\pi_T$ supplies top-$k$ log probabilities on the student's sampled tokens. Following on-policy distillation~\cite{agarwal2024,gu2024minillm}, the masked loss is
\begin{equation}
    \mathcal{L}_{\mathrm{OPD}}
    =\frac{\sum_t m_t\ell_t}{\sum_t m_t},
    \qquad
    \ell_t\approx \log\pi_\theta(r_t)-\log\pi_T(r_t).
    \label{eq:opd}
\end{equation}
No ground-truth label is available for $z$. The estimation problem is therefore to infer which perceptions are insufficient from downstream rewards and teacher probabilities alone.

\subsection{Separated Perception--Reasoning Rollout}
\label{sec:separated_rollout}

For each input, we first sample $a$ perceptions,
\begin{equation}
    z_i\sim\pi_p(\cdot\mid x),\qquad i=1,\ldots,a,
\end{equation}
and then sample $b$ independent reasoning continuations conditioned on each fixed perception,
\begin{equation}
    y_{ij}\sim\pi_r(\cdot\mid x,z_i),\qquad j=1,\ldots,b.
\end{equation}
The resulting $a\times b$ tree keeps the perception prefix shared within each group while allowing the reasoning to vary. All trajectories for an input remain one group for group-relative policy optimization~\cite{shao2024grpo,yu2025dapo}, so this reorganization does not change the reasoning reward.

\paragraph{Perception value.}
The downstream value of $z$ under the current reasoner is
\begin{equation}
    V(z)=\mathbb{E}_{y\sim\pi_r(\cdot\mid x,z)}[R(x,z,y)].
    \label{eq:perception_value}
\end{equation}
Its Monte Carlo estimate is the Perception Success Rate
\begin{equation}
    \mathrm{PSR}_i=\frac{1}{b}\sum_{j=1}^{b}R(x,z_i,y_{ij}).
    \label{eq:psr}
\end{equation}
For Bernoulli rewards, this estimator is unbiased and has variance $V(z_i)(1-V(z_i))/b$. Sharing $b$ continuations is therefore what converts a trajectory-level verifier into a lower-variance, per-perception statistic.

\subsection{Why Reward Alone Is Not Identifying}

Let $S(z)\in\{0,1\}$ denote the latent event that $z$ contains sufficient visual evidence. Define $q(z)=\Pr(S=1\mid z)$ and let $\rho(x)$ be the current reasoner's probability of success given sufficient evidence. If an insufficient perception rarely yields the correct answer, then
\begin{equation}
    V(z)
    =\rho(x)q(z)+\varepsilon(1-q(z))
    \approx \rho(x)q(z),
    \qquad \varepsilon\approx0.
    \label{eq:value_decomposition}
\end{equation}

\paragraph{Proposition 1 (non-identifiability of PSR).}
For any observed $v=\rho q\in(0,1)$ and any $q\in[v,1]$, choosing $\rho=v/q$ produces the same value $v$. Hence the reward sample identifies only the product of reasoning difficulty and perception sufficiency; no estimator based solely on $\{R_{ij}\}$ can distinguish a bad perception from a difficult reasoning problem.

\paragraph{Proof.}
The Bernoulli sampling distribution of Eq.~\eqref{eq:psr} depends on $(\rho,q)$ only through $V=\rho q$. Parameter pairs with equal products therefore induce identical observations, and no function of those observations can separate the pairs. \hfill$\square$

This proposition explains the failure mode of success-only weighting. On a hard prompt, $\rho$ can be small even when $q$ is close to one. Attenuating distillation whenever PSR is low then removes supervision from a perception that may already be adequate.

\subsection{Teacher Disagreement as a Second Witness}

Following reverse-KL distillation~\cite{gu2024minillm}, we measure student--teacher disagreement on the aware span with a detached estimate,
\begin{equation}
    \mathrm{KL}_i
    =\left\langle
      \log\pi^{\mathrm{old}}_\theta(t)-\log\pi_T(t)
    \right\rangle_{t\in z_i},
    \label{eq:aware_kl}
\end{equation}
using the same top-$k$ support and clamping convention as the distillation objective. The value is averaged across the $b$ trajectories that share $z_i$, clamped to be non-negative, and normalized to $\widetilde{\mathrm{KL}}_i\in[0,1]$.

Consider binary witnesses $A_i=\mathbb{1}[\mathrm{PSR}_i\text{ is low}]$ and $B_i=\mathbb{1}[\mathrm{KL}_i\text{ is large}]$. We assume that, conditional on the sufficiency state $S$, the residual randomness of these witnesses is independent; that an insufficient perception is more likely to disagree with a competent teacher; and that it is more likely to have low downstream success.

\paragraph{Proposition 2 (Bayesian combination of binary witnesses).}
Under these assumptions, the posterior log-odds of insufficiency satisfy
\begin{equation}
\begin{split}
    \operatorname{logit}\Pr(S=0\mid A,B)
    ={}&\operatorname{logit}\Pr(S=0)\\
    &+\log\frac{\Pr(A\mid S=0)}{\Pr(A\mid S=1)}\\
    &+\log\frac{\Pr(B\mid S=0)}{\Pr(B\mid S=1)}.
\end{split}
\label{eq:posterior_odds}
\end{equation}
Thus the two likelihood ratios multiply in odds space. This result applies exactly to calibrated binary tests. In practice we use the continuous, bounded score
\begin{equation}
    d_i=(1-\mathrm{PSR}_i)\widetilde{\mathrm{KL}}_i.
    \label{eq:deficiency}
\end{equation}

\paragraph{Proof.}
Conditional independence gives $\Pr(A,B\mid S)=\Pr(A\mid S)\Pr(B\mid S)$. Applying Bayes' rule, taking the ratio between $S=0$ and $S=1$, and then taking logarithms yields Eq.~\eqref{eq:posterior_odds}. \hfill$\square$

\paragraph{Scope of the assumptions.}
Conditional independence concerns the residual evidence in the two witnesses
after conditioning on perception sufficiency; it does not assert that reward
and model probabilities are marginally independent. Teacher competence is also
essential: disagreement is evidence of a correctable student error only when
the teacher is more reliable on the relevant visual evidence. Consequently,
PCD estimates \emph{teacher-correctable deficiency}, not semantic incorrectness
in an absolute sense. If teacher and student share the same visual failure,
their KL can be small and the error is intentionally not amplified.

Equation~\eqref{eq:deficiency} is not claimed to be a calibrated posterior probability without labeled insufficiency states. It is a conservative soft AND surrogate that preserves the desired boundary cases. High-PSR perceptions receive no boost regardless of KL. Low-PSR but teacher-aligned perceptions also receive no boost because the teacher offers little corrective information. Only low-PSR, high-KL perceptions receive strong additional supervision. This is the distinction that PSR alone cannot make.

\paragraph{Proposition 3 (uniqueness of the bilinear soft AND).}
Let $a=1-\mathrm{PSR}$ and $b=\widetilde{\mathrm{KL}}$. Among bilinear gates
$g(a,b)=c_0+c_1a+c_2b+c_3ab$, the conditions
\begin{equation}
    g(a,0)=0,\qquad g(0,b)=0,\qquad g(1,1)=1
    \label{eq:and_axioms}
\end{equation}
uniquely imply $g(a,b)=ab$.

\paragraph{Proof.}
The first condition in Eq.~\eqref{eq:and_axioms} gives $c_0=c_1=0$; the
second then gives $c_2=0$; normalization at $(1,1)$ gives $c_3=1$.
\hfill$\square$

The first two conditions encode the central modeling decision: neither low
success nor teacher disagreement alone should trigger correction. A normalized
additive gate $(a+b)/2$ violates both conditions. Proposition~3 establishes
multiplication within the simplest interaction model satisfying the desired
AND semantics; it does not exclude richer nonlinear gates.

\paragraph{Why not a learned gate?}
An MLP or attention module could represent richer interactions, but no labels
identify which failures are perceptual. Training such a gate from the same
trajectory reward would reintroduce the ambiguity of Proposition~1 and add a
new credit-assignment problem. PCD instead uses a parameter-free interaction
whose behavior is fixed before observing evaluation outcomes. Learned gates
remain a useful extension when perception-level supervision is available.

\subsection{Mean-Preserving Corrective Distillation}

PCD converts deficiency into a positive perception weight,
\begin{equation}
    \bar w_i=w_{\mathrm{base}}+\alpha d_i,
    \qquad
    w_i=\frac{\bar w_i}{\frac{1}{N}\sum_{k=1}^{N}\bar w_k},
    \label{eq:pcd_weight}
\end{equation}
where $N$ is the number of perceptions in the optimization batch. The second equality preserves the mean weight at one. The aware-span loss is
\begin{equation}
    \mathcal{L}_{\mathrm{aware}}
    =\frac{\sum_{i,j,t}w_i m^{\mathrm{aw}}_{ijt}\ell_{ijt}}
           {\sum_{i,j,t}m^{\mathrm{aw}}_{ijt}}.
    \label{eq:aware_loss}
\end{equation}

\paragraph{Budget interpretation.}
Equation~\eqref{eq:pcd_weight} is monotone in the deficiency score while satisfying $N^{-1}\sum_i w_i=1$. It therefore implements a controlled reallocation of a fixed supervision budget. We do not claim that this affine rule is the unique global optimum: a linear objective over an unconstrained simplex would concentrate all mass on one example. Instead, the positive base weight and normalization provide a bounded relaxation that preserves coverage of every perception. Without normalization, batches containing many deficient perceptions would also strengthen the overall imitation regularizer and confound selectivity with loss-scale changes.

\paragraph{Proposition 4 (optimal first-order reallocation direction).}
Let $\delta_i=w_i-1$ be a mean-preserving perturbation, so
$\sum_i\delta_i=0$, and suppose the local benefit of additional teacher
supervision is proportional to $d_i$. Among perturbations satisfying
$\lVert\delta\rVert_2\leq B$, the maximizer of the first-order benefit
$\sum_i d_i\delta_i$ is
\begin{equation}
    \delta_i^*=B\frac{d_i-\bar d}
    {\sqrt{\sum_k(d_k-\bar d)^2}},
    \qquad \bar d=\frac{1}{N}\sum_k d_k.
    \label{eq:optimal_reallocation}
\end{equation}
Moreover, Eq.~\eqref{eq:pcd_weight} gives exactly
\begin{equation}
    w_i-1=\frac{\alpha(d_i-\bar d)}
    {w_{\mathrm{base}}+\alpha\bar d},
    \label{eq:pcd_centered}
\end{equation}
and therefore follows this optimal direction while its base weight controls
the step magnitude and preserves coverage.

\paragraph{Proof.}
Projecting $d$ onto the zero-mean subspace gives $d-\bar d\mathbf{1}$.
For every feasible $\delta$, Cauchy--Schwarz yields
$\langle d,\delta\rangle=\langle d-\bar d\mathbf{1},\delta\rangle
\leq B\lVert d-\bar d\mathbf{1}\rVert_2$, with equality for
Eq.~\eqref{eq:optimal_reallocation}. Substituting the batch mean of
$\bar w_i=w_{\mathrm{base}}+\alpha d_i$ into
Eq.~\eqref{eq:pcd_weight} yields Eq.~\eqref{eq:pcd_centered}. \hfill$\square$

\subsection{Divided Objectives and Token-Level Composition}

We optimize the perception and reasoning spans with different objectives,
\begin{equation}
    \mathcal{L}
    =\lambda_{\mathrm{aw}}\mathcal{L}_{\mathrm{aware}}
    +\lambda_{\mathrm{cot}}\mathcal{L}^{\mathrm{DAPO}}_{\mathrm{cot}}.
    \label{eq:total_loss}
\end{equation}
PCD changes only $\mathcal{L}_{\mathrm{aware}}$; it does not alter group-relative advantages or the reasoning loss. If token-level visual focusing~\cite{huang2026vppo} supplies a saliency weight $s_{ijt}$, the aware-token contribution becomes $w_i s_{ijt}\ell_{ijt}$. PCD asks whether a perception is a correctable failure, while visual focusing asks which tokens carry image-dependent evidence.

\begin{algorithm}[t]
\caption{Perception-Correction Distillation}
\label{alg:pcd}
\begin{algorithmic}[1]
\REQUIRE Prompt $x$, student $(\pi_p,\pi_r)$, teacher $\pi_T$, rollout $(a,b)$
\FOR{$i=1,\ldots,a$}
    \STATE Sample $z_i$, then $b$ continuations $y_{ij}$ and rewards $R_{ij}$
    \STATE Compute $\mathrm{PSR}_i=b^{-1}\sum_jR_{ij}$ and aware-span $\mathrm{KL}_i$
    \STATE $d_i\leftarrow(1-\mathrm{PSR}_i)\widetilde{\mathrm{KL}}_i$;
    $\bar w_i\leftarrow w_{\rm base}+\alpha d_i$
\ENDFOR
\STATE Normalize $w_i\leftarrow\bar w_i/(N^{-1}\sum_k\bar w_k)$
\STATE Distill perception with $w_i$ and optimize reasoning with DAPO
\end{algorithmic}
\end{algorithm}

Thus PCD requires only grouped reductions and a trajectory-wise multiplier on
the existing aware-span loss. Stable perception identifiers preserve grouping
after batch reordering; missing masks or teacher probabilities set deficiency
to zero. Optional all-wrong exploration and complete failure handling are
described in the supplementary material.


\section{Experiments}
\label{sec:experiments}

We compare perception-specific correction with uniform on-policy
distillation and reward-only post-training in two transfer settings:
Qwen3-VL-8B$\rightarrow$2B and Qwen3-VL-32B$\rightarrow$8B. The comparison
includes the initial student, standard OPD, and PCD; the 2B block additionally
includes a same-size DAPO baseline.

\subsection{Experimental setup}
\label{sec:experimental_setup}

\paragraph{Settings and baselines.}
We study Qwen3-VL-8B-Instruct to Qwen3-VL-2B-Instruct and
Qwen3-VL-32B-Instruct to Qwen3-VL-8B-Instruct~\cite{bai2025qwen3vl}.
\textsc{Base} is the unmodified student, DAPO is the
same-size reward-optimization baseline, OPD applies token-level teacher
supervision uniformly, and PCD applies the separated rollout and deficiency
weighting from Section~\ref{sec:method}.

\paragraph{Training and evaluation.}
Training uses Geo3K~\cite{lu2021intergps}. The teacher returns top-$k$ probabilities with
$k=64$. PCD uses a separated $a=2,b=4$ rollout, giving eight completed
trajectories per prompt. The loss coefficients are
$\lambda_{\mathrm{aw}}=0.1$ and $\lambda_{\mathrm{cot}}=1.0$; PCD weights are
normalized to unit mean. Checkpoints are selected using held-out validation
performance.

We divide the eight evaluation datasets into three groups. \textbf{In-domain
(ID)} contains Geo3K, the dataset used for training. \textbf{Near-OOD
mathematical reasoning} contains MathVerse, MathVista, MATH-Vision, and
We-Math~\cite{zhang2024mathverse,lu2024mathvista,wang2024mathvision,qiao2025wemath},
which share the visual-mathematical task family without being the training
benchmark. \textbf{Out-of-domain (OOD) evaluation} contains LogicVista,
MMMU\_Pro, and MMStar~\cite{xiao2024logicvista,yue2025mmmupro,chen2024mmstar}. These labels denote
task-level proximity rather than example overlap. All datasets use the same prompting, decoding
configuration, and \textsc{Avg@8} evaluator with eight samples per question,
temperature $1.0$, and top-$p=1.0$.

For each question, Avg@8 is the arithmetic mean of the eight binary sample
outcomes; dataset accuracy then averages these per-question values, so every
question has equal weight regardless of answer length. The reported macro
average assigns equal weight to each of the eight datasets rather than to each
underlying question. This distinction matters because the benchmark sizes
differ substantially. We report percentages throughout and do not interpret a
single checkpoint difference as statistical significance.

\subsection{Main results}
\label{sec:main_results}

Table~\ref{tab:main_results} reports the eight-benchmark comparison. The first
block holds student capacity fixed at 2B and compares transfer from the 8B
teacher. The second block holds student capacity fixed at 8B and compares
uniform OPD and PCD using the 32B teacher.

\begin{table*}[t]
    \centering
    \footnotesize
    \setlength{\tabcolsep}{2.0pt}
    \renewcommand{\arraystretch}{1.03}
    \resizebox{\textwidth}{!}{%
    \begin{tabular}{cllccccccccc}
        \toprule
        \multirow{2}{*}{Student} &
        \multirow{2}{*}{Teacher} &
        \multirow{2}{*}{Method} &
        \multicolumn{1}{c}{ID} &
        \multicolumn{4}{c}{Near-OOD: Mathematical reasoning} &
        \multicolumn{3}{c}{OOD} &
        \multirow{2}{*}{Avg (macro)} \\
        \cmidrule(lr){4-4}
        \cmidrule(lr){5-8}
        \cmidrule(lr){9-11}
        & & &
        Geo3K &
        MathVerse &
        MathVista &
        MathVision &
        We-Math &
        LogicVista &
        MMMU\_Pro &
        MMStar &
        \\
        \midrule

        \multirow{4}{*}{\shortstack{Qwen3-VL\\2B-Instruct}}
        & -- & Base
        & 21.28 & 30.33 & 59.92 & 19.74 & 57.33
        & 45.23 & 45.06 & 57.77 & 42.08 \\

        & -- & DAPO
        & 37.83 & 37.03 & \textbf{63.49} & \textbf{25.25} & 60.50
        & 42.91 & 46.69 & 59.11 & 46.60 \\
        
        & 8B-Instruct & OPD
        & 30.82 & 33.14 & 61.55 & 20.19 & 59.78
        & \textbf{44.84} & 45.78 & \textbf{59.92} & 44.50 \\
        \cmidrule(lr){2-11}

        & 8B-Instruct & PCD
        & \textbf{42.12} & \textbf{37.72} & 62.45 & 24.01 & \textbf{62.44}
        & 42.89 & \textbf{47.59} & 59.06 & \textbf{47.28} \\

        \midrule

        \multirow{3}{*}{\shortstack{Qwen3-VL\\8B-Instruct}}
        & -- & Base
        & 41.99 & 45.15 & 73.91 & 34.54
        & 71.98 & \textbf{58.79} & 59.34 & 69.66
        & 56.92 \\

        & 32B-Instruct & OPD
        & 42.51 & 44.97 & 74.50 & 34.33
        & 72.19 & 58.51 & 59.05 & 69.44
        & 56.94 \\

        & 32B-Instruct & PCD
        & \textbf{59.65} & \textbf{50.10} & \textbf{75.79} & \textbf{39.88}
        & \textbf{73.92} & 58.57 & \textbf{61.45} & \textbf{70.43}
        & \textbf{61.22} \\

        \bottomrule
    \end{tabular}%
    }

    \caption{Avg@8 performance (\%). Geo3K is the ID training benchmark;
    Near-OOD contains visual-mathematical transfer tasks, and OOD contains
    more distant multimodal tasks. Avg is the unweighted eight-dataset macro
    mean; bold marks the best result per student size.}

    \label{tab:main_results}
\end{table*}

PCD obtains the highest macro average among the 2B models: 47.28, compared
with 46.60 for DAPO, 44.50 for OPD, and 42.08 for the initial student. It is
best on Geo3K, MathVerse, We-Math, and MMMU\_Pro; OPD leads on LogicVista and
MMStar, and DAPO on MathVista and MathVision. Thus PCD improves aggregate
performance without claiming universal per-task dominance.

In the 32B$\rightarrow$8B setting, PCD reaches a 61.22 macro average,
compared with 56.94 for OPD and 56.92 for the initial 8B student. It is best
on seven of eight benchmarks; Base remains slightly stronger on LogicVista
(58.79 vs. 58.57). The larger gain is consistent with a larger correctable
teacher--student perception gap, but does not establish causality.

\paragraph{Where does the gain occur?}
Relative to uniform OPD, the 2B PCD model improves Geo3K by $11.30$ points and
the four-dataset Near-OOD mean by $2.99$ points, while its three-dataset OOD
mean changes by $-0.33$. Thus the 2B macro gain is not evidence of uniform OOD
improvement; it is concentrated in visual mathematics, where the distilled
perception interface matches training most closely. The 32B$\rightarrow$8B
transfer is broader: PCD improves Geo3K by $17.14$, the Near-OOD mean by
$3.43$, and the OOD mean by $1.15$ points over OPD. This scale-dependent
pattern is consistent with stronger teachers providing more transferable
perceptual corrections, but model size and teacher quality are confounded in
this comparison.

\subsection{Mechanism Analysis: Does PCD Implement the AND Gate?}
\begin{figure}[t]
    \centering
    \includegraphics[width=0.80\columnwidth,trim=10bp 8bp 8bp 88bp,clip]{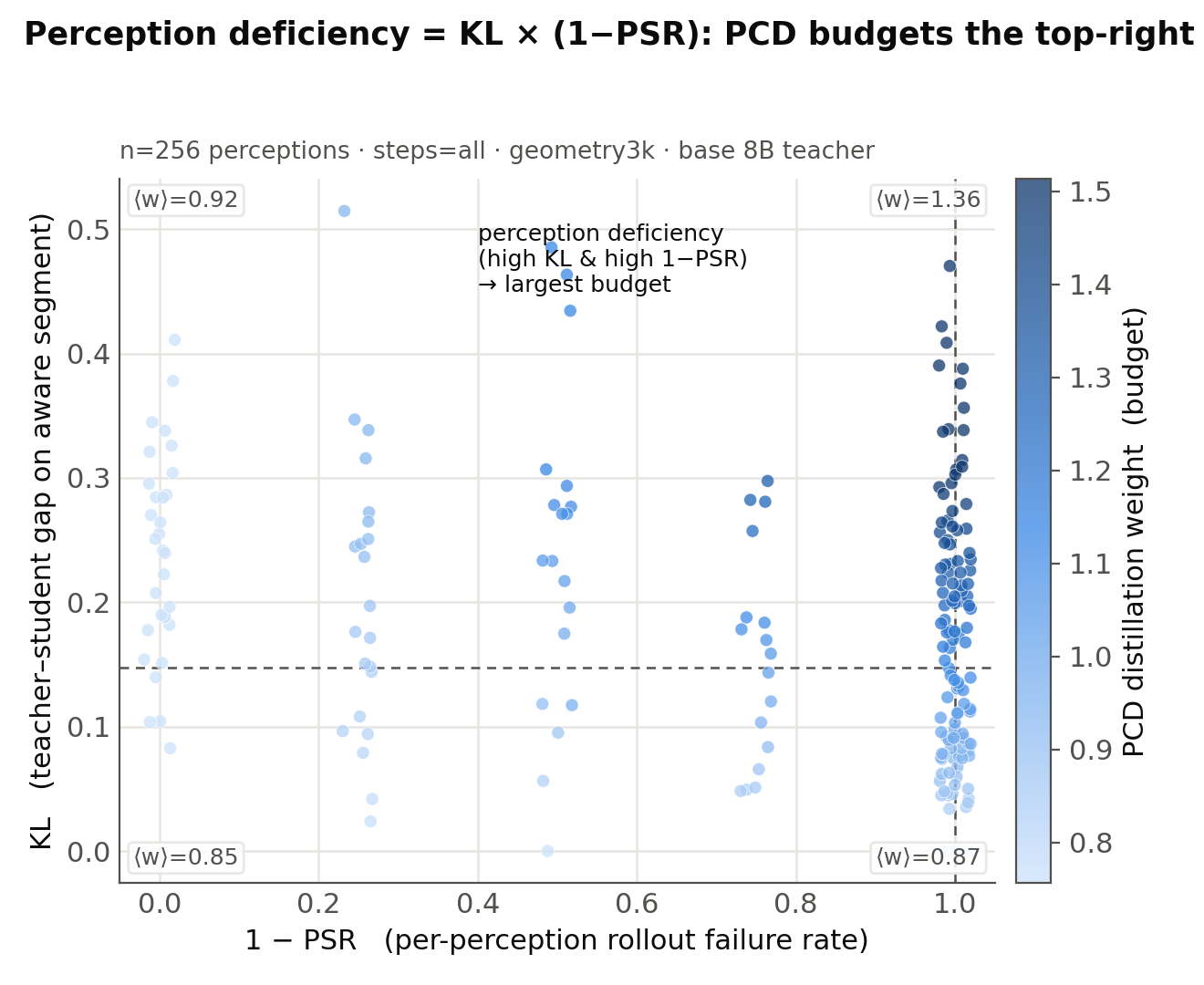}
    \caption{PCD training weights over 256 Geo3K perceptions concentrate in
    the high-failure, high-disagreement quadrant.}
    \label{fig:corr_kl_psr}
\end{figure}

Figure~\ref{fig:corr_kl_psr} maps 256 Geo3K perceptions by rollout failure
$(1-\mathrm{PSR})$ and teacher--student gap on the aware span; color is the
actual PCD weight used in training. The high-failure, high-gap quadrant has
mean weight $1.36$, versus $0.85$--$0.92$ elsewhere. Failure alone ($0.87$)
or disagreement alone ($0.92$) receives no comparable boost, matching the
multiplicative interaction. This validates allocation behavior rather than a
causal accuracy gain; Table~\ref{tab:component_ablation} supplies the
complementary endpoint evidence.

\subsection{Component Ablation}
\label{sec:component_ablation}

Table~\ref{tab:component_ablation} reports matched 2B ablations. Held-out avg
averages MMMU\_Pro, We-Math, MathVista, MathVerse, MathVision, and MMStar;
Geo3K is the training diagnostic and LogicVista was unavailable. Full combines
separated rollout, complete perception-span distillation, reasoning RL, and
PCD weighting. VPPO-Distill substitutes token-selective distillation; w/o PCD
uses uniform perception weights; w/o separated rollout samples both stages
jointly while retaining perception distillation and reasoning RL.

\begin{table*}[t]
    \centering
    \footnotesize
    \setlength{\tabcolsep}{2.6pt}
    \renewcommand{\arraystretch}{1.03}
    \resizebox{\textwidth}{!}{%
    \begin{tabular}{lccccccccc}
        \toprule
        Method & Geo3K (ID) & MMMU\_Pro & We-Math & MathVista & MathVerse &
        MathVision & MMStar & Held-out avg & $\Delta$ vs. full \\
        \midrule
        DAPO (pure RL)
        & 37.83 & 46.69 & 60.50 & 63.49 & 37.03 & 25.25 & 59.11 & 48.68 & \textbf{$-0.60$} \\
        Standard OPD
        & 30.82 & 45.78 & 59.78 & 61.55 & 33.14 & 20.19 & 59.92 & 46.73 & \textbf{$-2.55$} \\
        Full w/ VPPO-Distill
        & 42.12 & 47.59 & 62.44 & 62.45 & 37.72 & 24.01 & 59.06 & 48.88 & \textbf{$-0.40$} \\
        Full w/o PCD weight
        & 37.90 & 45.73 & 58.76 & 61.37 & 34.81 & 23.77 & 57.94 & 47.06 & \textbf{$-2.22$} \\
        Full w/o separated rollout
        & 38.50 & 46.26 & 61.32 & 63.58 & 36.26 & 23.31 & 59.69 & 48.40 & \textbf{$-0.88$} \\
        \textbf{Full (ours)}
        & \textbf{43.12} & \textbf{47.15} & \textbf{63.68} & \textbf{63.16} &
        \textbf{39.19} & \textbf{22.94} & \textbf{59.59} & \textbf{49.28} & \textbf{--} \\
        \bottomrule
    \end{tabular}%
    }
    \caption{Matched 2B component ablation. Held-out avg is the unweighted mean
    over the six listed transfer datasets and excludes Geo3K. Full combines
    separated rollout, perception distillation, reasoning RL, and PCD
    weighting; bold identifies the full configuration.}
    \label{tab:component_ablation}
\end{table*}

Full achieves the best held-out average (49.28). Uniformizing its PCD weight
causes the largest internal drop ($-2.22$), linking the allocation pattern in
Figure~\ref{fig:corr_kl_psr} to endpoint performance. Removing separated
rollout costs $0.88$, consistent with weaker perception-level PSR and credit
assignment when continuations do not share a fixed perception. VPPO-Distill
costs $0.40$, suggesting that selecting only highly visual tokens can discard
contextual or implicitly visual perception tokens. OPD trails by $2.55$ and
DAPO by $0.60$; DAPO remains stronger on MathVision, so the evidence is
aggregate rather than universal per benchmark.

\paragraph{Fixed-budget interpretation.}
Because PCD normalizes the batch-average perception weight to one, its gain
cannot be attributed to a larger global teacher-loss coefficient: deficient
perceptions receive more supervision only by reducing it elsewhere. The
w/o-PCD comparison is therefore a controlled allocation test---it retains
the separated rollout and divided objectives, but loses $2.22$ points when
all perceptions are weighted uniformly. VPPO-Distill operates at a different
granularity and loses $0.40$: selecting only highly visual tokens may omit
connective or implicitly visual tokens needed to express a coherent
perception. These results motivate PCD as a trajectory-level allocator rather
than another token mask, while leaving open whether a differently calibrated
token selector could compose more effectively.

\subsection{Limitations and Threats to Validity}

\paragraph{Attribution.}
The ablation isolates PCD weighting, separated rollout, and VPPO-Distill, but
lacks PSR-only, KL-only, and additive-fusion runs. Figure~\ref{fig:corr_kl_psr}
verifies the interaction and w/o-PCD verifies adaptive weighting; matched
single-witness runs are still needed to attribute the gain to multiplication.

\paragraph{Uncertainty and non-stationarity.}
Each method uses one selected checkpoint. PSR is policy-dependent, and larger
$b$ reduces its conditional variance while exploring fewer perceptions under a
fixed budget. Multi-seed intervals and sensitivity to $(a,b)$ are therefore
needed before treating small differences as stable.

\paragraph{Teacher and KL calibration.}
Disagreement is useful only for a perceptually stronger teacher. Shared errors
produce low KL, whereas a confidently wrong teacher can induce harmful
correction. Because the normalization threshold controls KL saturation, it
should be calibrated from training-trace quantiles for each teacher size.

\paragraph{Scope and reproducibility.}
ID/Near-OOD/OOD are task descriptors rather than formal distances, and results
cover one model family and training domain. Full implementation and failure
handling are documented in the supplementary material.

\section{Conclusion}
\label{sec:conclusion}

PCD recasts perception distillation as credit assignment. Combining failure
with teacher disagreement targets correctable perceptions while preserving
the mean supervision budget. It improves OPD by 2.78 and 4.28 points for 2B
and 8B students, with mechanism and ablation evidence supporting the proposed
allocation. Multi-seed and single-witness controls remain future tests.

\bibliographystyle{plainnat}
\bibliography{aaai2027}

@inproceedings{kim2016sequence,
  author    = {Kim, Yoon and Rush, Alexander M.},
  title     = {Sequence-Level Knowledge Distillation},
  booktitle = {EMNLP},
  pages     = {1317--1327},
  year      = {2016},
  doi       = {10.18653/v1/D16-1139},
  url       = {https://aclanthology.org/D16-1139/}
}

@inproceedings{gu2024minillm,
  author    = {Gu, Yuxian and Dong, Li and Wei, Furu and Huang, Minlie},
  title     = {{MiniLLM}: Knowledge Distillation of Large Language Models},
  booktitle = {ICLR},
  year      = {2024},
  url       = {https://openreview.net/forum?id=5h0qf7IBZZ}
}

@inproceedings{agarwal2024,
  author    = {Agarwal, Rishabh and Vieillard, Nino and Zhou, Yongchao and others},
  title     = {On-Policy Distillation of Language Models: Learning from Self-Generated Mistakes},
  booktitle = {ICLR},
  year      = {2024},
  url       = {https://openreview.net/forum?id=3zKtaqxLhW},
  eprint    = {2306.13649},
  archivePrefix = {arXiv}
}

@article{lu2025onpolicy,
  author  = {Lu, Kevin and {Thinking Machines Lab}},
  title   = {On-Policy Distillation},
  journal = {Thinking Machines Lab: Connectionism},
  year    = {2025},
  doi     = {10.64434/tml.20251026},
  url     = {https://thinkingmachines.ai/blog/on-policy-distillation/}
}

@inproceedings{huang2026vppo,
  author    = {Huang, Siyuan and Qu, Xiaoye and Li, Yafu and others},
  title     = {Spotlight on Token Perception for Multimodal Reinforcement Learning},
  booktitle = {ICLR},
  year      = {2026},
  doi       = {10.48550/arXiv.2510.09285},
  url       = {https://arxiv.org/abs/2510.09285},
  eprint    = {2510.09285},
  archivePrefix = {arXiv}
}

@article{shao2024grpo,
  author  = {Shao, Zhihong and Wang, Peiyi and Zhu, Qihao and others},
  title   = {{DeepSeekMath}: Pushing the Limits of Mathematical Reasoning in Open Language Models},
  journal = {arXiv preprint arXiv:2402.03300},
  year    = {2024},
  doi     = {10.48550/arXiv.2402.03300},
  url     = {https://arxiv.org/abs/2402.03300}
}

@article{yu2025dapo,
  author  = {Yu, Qiying and Zhang, Zheng and Zhu, Ruofei and others},
  title   = {{DAPO}: An Open-Source LLM Reinforcement Learning System at Scale},
  journal = {arXiv:2503.14476},
  year    = {2025},
  doi     = {10.48550/arXiv.2503.14476},
  url     = {https://arxiv.org/abs/2503.14476}
}

@inproceedings{lightman2024verify,
  author    = {Lightman, Hunter and Kosaraju, Vineet and Burda, Yura and others},
  title     = {Let's Verify Step by Step},
  booktitle = {ICLR},
  year      = {2024},
  url       = {https://openreview.net/forum?id=v8L0pN6EOi},
  eprint    = {2305.20050},
  archivePrefix = {arXiv}
}

@article{hinton2015distilling,
  author  = {Hinton, Geoffrey and Vinyals, Oriol and Dean, Jeff},
  title   = {Distilling the Knowledge in a Neural Network},
  journal = {arXiv preprint arXiv:1503.02531},
  year    = {2015},
  doi     = {10.48550/arXiv.1503.02531},
  url     = {https://arxiv.org/abs/1503.02531}
}

@inproceedings{beyer2022patient,
  author    = {Beyer, Lucas and Zhai, Xiaohua and Royer, Am{\'e}lie and others},
  title     = {Knowledge Distillation: A Good Teacher Is Patient and Consistent},
  booktitle = {Proceedings of the IEEE/CVF Conference on Computer Vision and Pattern Recognition},
  pages     = {10925--10934},
  year      = {2022},
  url       = {https://openaccess.thecvf.com/content/CVPR2022/html/Beyer_Knowledge_Distillation_A_Good_Teacher_Is_Patient_and_Consistent_CVPR_2022_paper.html}
}

@article{schulman2017ppo,
  author  = {Schulman, John and Wolski, Filip and Dhariwal, Prafulla and Radford, Alec and Klimov, Oleg},
  title   = {Proximal Policy Optimization Algorithms},
  journal = {arXiv preprint arXiv:1707.06347},
  year    = {2017},
  doi     = {10.48550/arXiv.1707.06347},
  url     = {https://arxiv.org/abs/1707.06347}
}

@inproceedings{wang2024mathshepherd,
  author    = {Wang, Peiyi and Li, Lei and Shao, Zhihong and others},
  title     = {{Math-Shepherd}: Verify and Reinforce {LLM}s Step-by-Step without Human Annotations},
  booktitle = {ACL},
  pages     = {9426--9439},
  year      = {2024},
  doi       = {10.18653/v1/2024.acl-long.510},
  url       = {https://aclanthology.org/2024.acl-long.510/}
}

@inproceedings{arjona2019rudder,
  author    = {Arjona-Medina, Jose A. and Gillhofer, Michael and Widrich, Michael and Unterthiner, Thomas and Brandstetter, Johannes and Hochreiter, Sepp},
  title     = {{RUDDER}: Return Decomposition for Delayed Rewards},
  booktitle = {NeurIPS},
  volume    = {32},
  year      = {2019},
  url       = {https://proceedings.neurips.cc/paper/2019/hash/16105fb9cc614fc29e1bda00dab60d41-Abstract.html}
}

@article{deepseekai2025r1,
  author  = {{DeepSeek-AI}},
  title   = {{DeepSeek-R1}: Incentivizing Reasoning Capability in {LLM}s via Reinforcement Learning},
  journal = {arXiv preprint arXiv:2501.12948},
  year    = {2025},
  doi     = {10.48550/arXiv.2501.12948},
  url     = {https://arxiv.org/abs/2501.12948}
}

@article{wang2024qwen2vl,
  author  = {Wang, Peng and Bai, Shuai and Tan, Sinan and others},
  title   = {{Qwen2-VL}: Enhancing Vision-Language Model's Perception of the World at Any Resolution},
  journal = {arXiv preprint arXiv:2409.12191},
  year    = {2024},
  doi     = {10.48550/arXiv.2409.12191},
  url     = {https://arxiv.org/abs/2409.12191}
}

@inproceedings{shao2024visualcot,
  author    = {Shao, Hao and Qian, Shengju and Xiao, Han and others},
  title     = {Visual {CoT}: Advancing Multi-Modal Language Models with a Comprehensive Dataset and Benchmark for Chain-of-Thought Reasoning},
  booktitle = {NeurIPS},
  volume    = {37},
  year      = {2024},
  doi       = {10.52202/079017-0275},
  url       = {https://proceedings.neurips.cc/paper_files/paper/2024/hash/0ff38d72a2e0aa6dbe42de83a17b2223-Abstract-Datasets_and_Benchmarks_Track.html}
}

@article{he2024multimodal,
  author  = {He, Liqi and Li, Zuchao and Cai, Xiantao and Wang, Ping},
  title   = {Multi-Modal Latent Space Learning for Chain-of-Thought Reasoning in Language Models},
  journal = {AAAI},
  volume  = {38},
  number  = {16},
  pages   = {18180--18187},
  year    = {2024},
  doi     = {10.1609/aaai.v38i16.29776},
  url     = {https://ojs.aaai.org/index.php/AAAI/article/view/29776}
}

@article{chen2024vctp,
  author  = {Chen, Zhenfang and Zhou, Qinhong and Shen, Yikang and others},
  title   = {Visual Chain-of-Thought Prompting for Knowledge-Based Visual Reasoning},
  journal = {AAAI},
  volume  = {38},
  number  = {2},
  pages   = {1254--1262},
  year    = {2024},
  doi     = {10.1609/aaai.v38i2.27888},
  url     = {https://ojs.aaai.org/index.php/AAAI/article/view/27888}
}

@article{wang2024tsciq,
  author  = {Wang, Lei and Hu, Yi and He, Jiabang and others},
  title   = {{T-SciQ}: Teaching Multimodal Chain-of-Thought Reasoning via Large Language Model Signals for Science Question Answering},
  journal = {AAAI},
  volume  = {38},
  number  = {17},
  pages   = {19162--19170},
  year    = {2024},
  doi     = {10.1609/aaai.v38i17.29884},
  url     = {https://ojs.aaai.org/index.php/AAAI/article/view/29884}
}

@article{cheng2025comt,
  author  = {Cheng, Zihui and Chen, Qiguang and Zhang, Jin and others},
  title   = {{CoMT}: A Novel Benchmark for Chain of Multi-Modal Thought on Large Vision-Language Models},
  journal = {AAAI},
  volume  = {39},
  number  = {22},
  pages   = {23678--23686},
  year    = {2025},
  doi     = {10.1609/aaai.v39i22.34538},
  url     = {https://ojs.aaai.org/index.php/AAAI/article/view/34538}
}

@techreport{bai2025qwen3vl,
  author      = {Bai, Shuai and Cai, Yuxuan and Chen, Ruizhe and others},
  title       = {{Qwen3-VL} Technical Report},
  institution = {Qwen Team},
  type        = {Technical Report},
  number      = {arXiv:2511.21631},
  year        = {2025},
  doi         = {10.48550/arXiv.2511.21631},
  url         = {https://arxiv.org/abs/2511.21631}
}

@techreport{bai2025qwen25vl,
  author      = {Bai, Shuai and Chen, Keqin and Liu, Xuejing and others},
  title       = {{Qwen2.5-VL} Technical Report},
  institution = {Qwen Team},
  type        = {Technical Report},
  number      = {arXiv:2502.13923},
  year        = {2025},
  doi         = {10.48550/arXiv.2502.13923},
  url         = {https://arxiv.org/abs/2502.13923}
}

@techreport{chen2024internvl25,
  author      = {Chen, Zhe and Wang, Weiyun and Cao, Yue and others},
  title       = {Expanding Performance Boundaries of Open-Source Multimodal
                 Models with Model, Data, and Test-Time Scaling},
  institution = {OpenGVLab},
  type        = {Technical Report},
  number      = {arXiv:2412.05271},
  year        = {2024},
  doi         = {10.48550/arXiv.2412.05271},
  url         = {https://arxiv.org/abs/2412.05271}
}

@article{wang2025visualprm,
  author  = {Wang, Weiyun and Gao, Zhangwei and Chen, Lianjie and others},
  title   = {{VisualPRM}: An Effective Process Reward Model for Multimodal Reasoning},
  journal = {arXiv preprint arXiv:2503.10291},
  year    = {2025},
  doi     = {10.48550/arXiv.2503.10291},
  url     = {https://arxiv.org/abs/2503.10291}
}

@inproceedings{lu2021intergps,
  author    = {Lu, Pan and Gong, Ran and Jiang, Shibiao and others},
  title     = {{Inter-GPS}: Interpretable Geometry Problem Solving with Formal
               Language and Symbolic Reasoning},
  booktitle = {ACL-IJCNLP},
  pages     = {6774--6786},
  year      = {2021},
  doi       = {10.18653/v1/2021.acl-long.528},
  url       = {https://aclanthology.org/2021.acl-long.528/}
}

@inproceedings{lu2024mathvista,
  author    = {Lu, Pan and Bansal, Hritik and Xia, Tony and others},
  title     = {{MathVista}: Evaluating Mathematical Reasoning of Foundation
               Models in Visual Contexts},
  booktitle = {ICLR},
  year      = {2024},
  url       = {https://openreview.net/forum?id=KUNzEQMWU7},
  eprint    = {2310.02255},
  archivePrefix = {arXiv}
}

@inproceedings{zhang2024mathverse,
  author    = {Zhang, Renrui and Jiang, Dongzhi and Zhang, Yichi and others},
  title     = {{MathVerse}: Does Your Multi-modal {LLM} Truly See the Diagrams
               in Visual Math Problems?},
  booktitle = {ECCV},
  year      = {2024},
  doi       = {10.48550/arXiv.2403.14624},
  url       = {https://arxiv.org/abs/2403.14624}
}

@inproceedings{wang2024mathvision,
  author    = {Wang, Ke and Pan, Junting and Shi, Weikang and others},
  title     = {Measuring Multimodal Mathematical Reasoning with
               {MATH-Vision} Dataset},
  booktitle = {NeurIPS},
  volume    = {37},
  year      = {2024},
  doi       = {10.48550/arXiv.2402.14804},
  url       = {https://arxiv.org/abs/2402.14804}
}

@inproceedings{qiao2025wemath,
  author    = {Qiao, Runqi and Tan, Qiuna and Dong, Guanting and others},
  title     = {{We-Math}: Does Your Large Multimodal Model Achieve Human-like
               Mathematical Reasoning?},
  booktitle = {ACL},
  pages     = {20023--20070},
  year      = {2025},
  doi       = {10.18653/v1/2025.acl-long.983},
  url       = {https://aclanthology.org/2025.acl-long.983/}
}

@article{xiao2024logicvista,
  author  = {Xiao, Yijia and Sun, Edward and Liu, Tianyu and Wang, Wei},
  title   = {{LogicVista}: Multimodal {LLM} Logical Reasoning Benchmark in
             Visual Contexts},
  journal = {arXiv preprint arXiv:2407.04973},
  year    = {2024},
  doi     = {10.48550/arXiv.2407.04973},
  url     = {https://arxiv.org/abs/2407.04973}
}

@inproceedings{yue2025mmmupro,
  author    = {Yue, Xiang and Zheng, Tianyu and Ni, Yuansheng and others},
  title     = {{MMMU-Pro}: A More Robust Multi-discipline Multimodal
               Understanding Benchmark},
  booktitle = {ACL},
  pages     = {15134--15186},
  year      = {2025},
  doi       = {10.18653/v1/2025.acl-long.736},
  url       = {https://aclanthology.org/2025.acl-long.736/}
}

@inproceedings{chen2024mmstar,
  author    = {Chen, Lin and Li, Jinsong and Dong, Xiaoyi and others},
  title     = {Are We on the Right Way for Evaluating Large Vision-Language
               Models?},
  booktitle = {NeurIPS},
  volume    = {37},
  year      = {2024},
  doi       = {10.48550/arXiv.2403.20330},
  url       = {https://arxiv.org/abs/2403.20330}
}

@inproceedings{furlanello2018born,
  author    = {Furlanello, Tommaso and Lipton, Zachary and Tschannen, Michael
               and Itti, Laurent and Anandkumar, Anima},
  title     = {Born Again Neural Networks},
  booktitle = {ICML},
  volume    = {80},
  pages     = {1607--1616},
  year      = {2018},
  publisher = {PMLR},
  url       = {https://proceedings.mlr.press/v80/furlanello18a.html}
}

@inproceedings{hsieh2023distilling,
  author    = {Hsieh, Cheng-Yu and Li, Chun-Liang and Yeh, Chih-Kuan and
               Nakhost, Hootan and Fujii, Yasuhisa and Ratner, Alexander and
               Krishna, Ranjay and Lee, Chen-Yu and Pfister, Tomas},
  title     = {Distilling Step-by-Step! Outperforming Larger Language Models
               with Less Training Data and Smaller Model Sizes},
  booktitle = {Findings of ACL},
  pages     = {8003--8017},
  year      = {2023},
  doi       = {10.18653/v1/2023.findings-acl.507},
  url       = {https://aclanthology.org/2023.findings-acl.507/}
}

@inproceedings{rusu2016policy,
  author    = {Rusu, Andrei A. and Colmenarejo, Sergio Gomez and Gulcehre,
               Caglar and Desjardins, Guillaume and Kirkpatrick, James and
               Pascanu, Razvan and Mnih, Volodymyr and Kavukcuoglu, Koray and
               Hadsell, Raia},
  title     = {Policy Distillation},
  booktitle = {ICLR},
  year      = {2016},
  eprint    = {1511.06295},
  archivePrefix = {arXiv},
  url       = {https://arxiv.org/abs/1511.06295}
}

@inproceedings{czarnecki2019distilling,
  author    = {Czarnecki, Wojciech M. and Pascanu, Razvan and Osindero, Simon
               and Jayakumar, Siddhant M. and Swirszcz, Grzegorz and
               Jaderberg, Max},
  title     = {Distilling Policy Distillation},
  booktitle = {AISTATS},
  volume    = {89},
  pages     = {1331--1340},
  year      = {2019},
  publisher = {PMLR},
  url       = {https://proceedings.mlr.press/v89/czarnecki19a.html}
}

@article{zhao2024multihead,
  author  = {Zhao, Tianyang and Singh, Kunwar Yashraj and Appalaraju, Srikar
             and Tang, Peng and Mahadevan, Vijay and Manmatha, R. and Wu,
             Ying Nian},
  title   = {No Head Left Behind: Multi-Head Alignment Distillation for
             Transformers},
  journal = {AAAI},
  volume  = {38},
  number  = {7},
  pages   = {7514--7524},
  year    = {2024},
  doi     = {10.1609/aaai.v38i7.28583},
  url     = {https://ojs.aaai.org/index.php/AAAI/article/view/28583}
}

@article{yang2025multimodalkd,
  author  = {Yang, Shuo and Luo, Siwen and Han, Soyeon Caren},
  title   = {Multimodal Commonsense Knowledge Distillation for Visual
             Question Answering},
  journal = {AAAI},
  volume  = {39},
  number  = {28},
  pages   = {29545--29547},
  year    = {2025},
  doi     = {10.1609/aaai.v39i28.35320},
  url     = {https://ojs.aaai.org/index.php/AAAI/article/view/35320}
}

@article{feng2026emkd,
  author  = {Feng, Ze and Yang, Sen and Duan, Boqiang and Yang, Wankou and
             Wang, Jingdong},
  title   = {{EM-KD}: Distilling Efficient Multimodal Large Language Model
             with Unbalanced Vision Tokens},
  journal = {AAAI},
  volume  = {40},
  number  = {25},
  year    = {2026},
  doi     = {10.1609/aaai.v40i25.39254},
  url     = {https://ojs.aaai.org/index.php/AAAI/article/view/39254}
}

@article{mondal2024kamcot,
  author  = {Mondal, Debjyoti and Modi, Suraj and Panda, Subhadarshi and
             Singh, Rituraj and Rao, Godawari Sudhakar},
  title   = {{KAM-CoT}: Knowledge Augmented Multimodal Chain-of-Thoughts
             Reasoning},
  journal = {AAAI},
  volume  = {38},
  number  = {17},
  pages   = {18798--18806},
  year    = {2024},
  doi     = {10.1609/aaai.v38i17.29844},
  url     = {https://ojs.aaai.org/index.php/AAAI/article/view/29844}
}

@article{chen2024stas,
  author  = {Chen, Sirui and Zhang, Zhaowei and Yang, Yaodong and Du, Yali},
  title   = {{STAS}: Spatial-Temporal Return Decomposition for Solving Sparse
             Rewards Problems in Multi-Agent Reinforcement Learning},
  journal = {AAAI},
  volume  = {38},
  number  = {16},
  pages   = {17337--17345},
  year    = {2024},
  doi     = {10.1609/aaai.v38i16.29681},
  url     = {https://ojs.aaai.org/index.php/AAAI/article/view/29681}
}

@article{qu2025latent,
  author  = {Qu, Yun and Jiang, Yuhang and Wang, Boyuan and Mao, Yixiu and
             Wang, Cheems and Liu, Chang and Ji, Xiangyang},
  title   = {Latent Reward: {LLM}-Empowered Credit Assignment in Episodic
             Reinforcement Learning},
  journal = {AAAI},
  volume  = {39},
  number  = {19},
  pages   = {20095--20103},
  year    = {2025},
  doi     = {10.1609/aaai.v39i19.34213},
  url     = {https://ojs.aaai.org/index.php/AAAI/article/view/34213}
}

@article{zhou2026visualprm,
  author  = {Zhou, Yujin and Wen, Pengcheng and Chen, Jiale and Yin, Boqin and
             Zhu, Han and Ji, Jiaming and Dai, Juntao and Chan, Chi-Min and
             Han, Sirui},
  title   = {What, Whether and How? Unveiling Process Reward Models for
             Thinking with Images Reasoning},
  journal = {AAAI},
  volume  = {40},
  number  = {34},
  pages   = {29071--29079},
  year    = {2026},
  doi     = {10.1609/aaai.v40i34.40144},
  url     = {https://ojs.aaai.org/index.php/AAAI/article/view/40144}
}

@article{yuan2026visionopd,
  author  = {Yuan, Qianhao and Lou, Jie and Yu, Xing and Lin, Hongyu and
             Sun, Le and Han, Xianpei and Lu, Yaojie},
  title   = {{Vision-OPD}: Learning to See Fine Details for Multimodal {LLM}s
             via On-Policy Self-Distillation},
  journal = {arXiv preprint arXiv:2605.18740},
  year    = {2026},
  doi     = {10.48550/arXiv.2605.18740},
  url     = {https://arxiv.org/abs/2605.18740}
}

\appendix
\section{Supplementary Overview and Terminology}

This supplement provides the implementation details, complete experimental
protocol, qualitative examples, and full transfer
results omitted from the main paper for space. We follow the notation and
terminology of the main paper throughout. In particular, a response is divided
into a perception span $z$ enclosed by \texttt{<aware>} tags and a reasoning
span $y$ enclosed by \texttt{<cot>} tags. Perception-Correction Distillation
(PCD) denotes the complete method: separated perception--reasoning rollout,
the multiplicative deficiency score, and mean-preserving perception
distillation. Standard on-policy distillation (OPD) applies teacher
supervision uniformly. DAPO denotes the reward-only policy-optimization
baseline. Perception Success Rate (PSR) is the mean verifier reward over the
$b$ reasoning continuations that share one sampled perception. We use
$\widetilde{\mathrm{KL}}$ for the normalized student--teacher disagreement on
the perception span.

Unless explicitly marked as an optional extension, all references to PCD in
this supplement use the default configuration evaluated in the main paper:
$a=2$ perceptions, $b=4$ continuations per perception, the multiplicative gate
$(1-\mathrm{PSR})\widetilde{\mathrm{KL}}$, and unit-mean normalized weights.
We use \textsc{Avg@$n$} for the mean of $n$ independently sampled binary
outcomes per question, matching the definition in the main paper.

\section{Implementation Details}
\label{sec:supp_implementation}

\subsection{Separated Rollout and Group Bookkeeping}

For each image--question pair $x$, the student first samples $a$ perception
spans $z_i$. It then samples $b$ independent reasoning spans $y_{ij}$ while
holding each $z_i$ fixed. Every perception receives a stable identifier before
continuation generation. The identifier is copied to its $b$ continuations
and retained through padding, shuffling, and mini-batch reordering. PSR and
teacher disagreement are therefore reduced by identifier rather than by
temporary tensor position. All $ab$ completed trajectories from the same
input remain in one reward group for DAPO, so separated sampling changes the
credit-assignment unit without changing the reasoning advantage group.

The \texttt{<aware>} and \texttt{<cot>} masks are reconstructed from decoded
span delimiters. Teacher disagreement is computed from detached old-policy
and teacher log probabilities on valid perception-span tokens, averaged first
within each trajectory and then across the continuations sharing a perception.
The resulting KL estimate is clamped to be non-negative and normalized to
$\widetilde{\mathrm{KL}}\in[0,1]$. The unnormalized perception weight is
$\bar w_i=w_{\mathrm{base}}+\alpha(1-\mathrm{PSR}_i)
\widetilde{\mathrm{KL}}_i$; division by the batch mean of $\bar w_i$ makes the
final weights average to one.

\subsection{Failure Handling}

Malformed responses remain eligible for task reward and reasoning-policy
optimization, but tokens outside a valid \texttt{<aware>} span are excluded
from perception distillation. If a trajectory has an empty perception span,
missing teacher scores, or no valid shared tokens, its deficiency is set to
zero. This fallback recovers the base span-restricted distillation weight and
prevents invalid values from affecting other perceptions during batch
normalization. PCD requires no learned scorer and no additional student
forward pass; beyond the teacher scores already used by OPD, it adds only
grouped reductions and one trajectory-wise weight multiplication.

\begin{algorithm}[t]
\caption{Perception-Correction Distillation}
\label{alg:supp_pcd}
\begin{algorithmic}[1]
\REQUIRE Input $x$, student $(\pi_p,\pi_r)$, teacher $\pi_T$, rollout $(a,b)$
\FOR{$i=1,\ldots,a$}
    \STATE Sample $z_i\sim\pi_p(\cdot\mid x)$
    \FOR{$j=1,\ldots,b$}
        \STATE Sample $y_{ij}\sim\pi_r(\cdot\mid x,z_i)$ and evaluate $R_{ij}$
    \ENDFOR
    \STATE Compute $\mathrm{PSR}_i$, perception-span $\mathrm{KL}_i$, and
    $d_i=(1-\mathrm{PSR}_i)\widetilde{\mathrm{KL}}_i$
    \STATE $\bar w_i\leftarrow w_{\rm base}+\alpha d_i$
\ENDFOR
\STATE Let $\mathcal{B}_p$ contain all valid perceptions in the optimization batch
\STATE $\mu_w\leftarrow|\mathcal{B}_p|^{-1}
\sum_{k\in\mathcal{B}_p}\bar w_k$
\STATE Normalize $w_i\leftarrow\bar w_i/\mu_w$
\STATE Optimize perception distillation with $w_i$ and reasoning with RL
\end{algorithmic}
\end{algorithm}

\IfFileExists{Figures/gate_comparison.png}{%
\begin{figure}[t]
    \centering
    \includegraphics[width=\columnwidth]{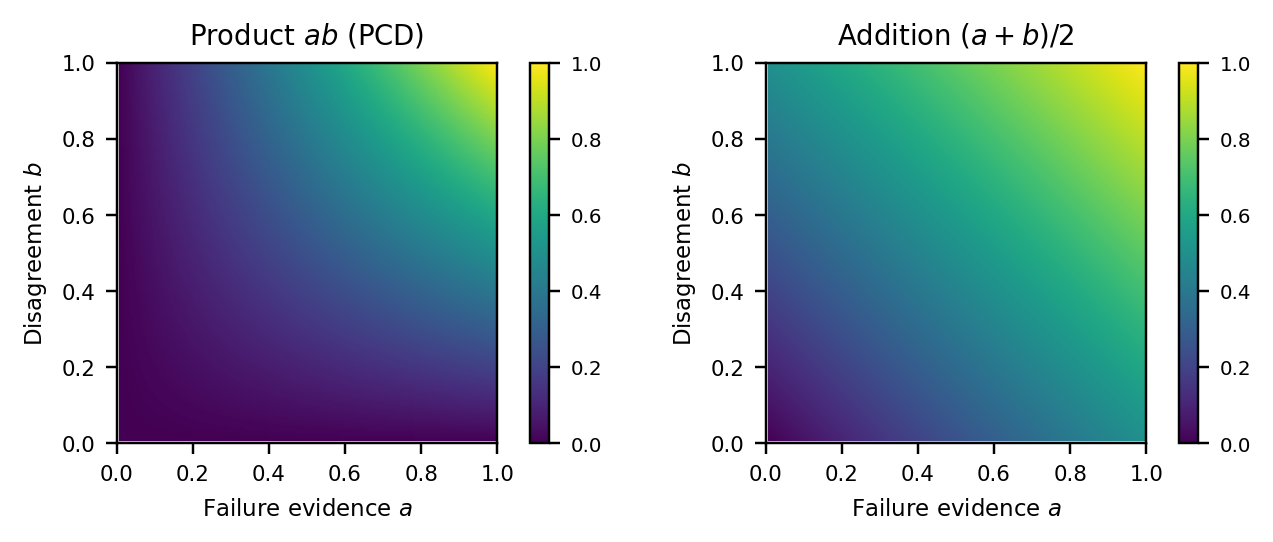}
    \caption{Multiplication is zero on both single-witness axes, whereas
    normalized addition assigns a positive correction when only one witness is
    present.}
    \label{fig:supp_gate_comparison}
\end{figure}
}{}

\section{Experimental Protocol}
\label{sec:supp_protocol}

\subsection{Models, Data, and Compared Methods}

We evaluate two teacher--student transfers from the Qwen3-VL-Instruct family:
8B$\rightarrow$2B and 32B$\rightarrow$8B. Training uses Geo3K, and checkpoint
selection uses a held-out validation split. \textsc{Base} is the unmodified
student. DAPO applies reward-only policy optimization with the same student
and rollout budget. Standard OPD adds uniform teacher supervision. PCD uses
the same optimization recipe as OPD but replaces joint sampling with the
separated rollout and replaces uniform perception weights with the
mean-preserving deficiency weights described in the main paper.

The primary evaluation comprises eight benchmarks. Geo3K is in-domain (ID).
MathVerse, MathVista, MATH-Vision, and We-Math are near-out-of-domain
(Near-OOD) visual-mathematical tasks. LogicVista, MMMU\_Pro, and MMStar form
the more distant out-of-domain (OOD) group. These group names indicate task
similarity, not example-level overlap.

\subsection{Rollout Generation and Optimization}

Rollouts are generated with the \texttt{separated\_two\_stage\_agent}. Stage
one samples a perception until \texttt{</aware>} or 512 generated tokens;
stage two conditions on that fixed prefix and samples the \texttt{<cot>}
reasoning continuation. The default split is $a=2$ perceptions and $b=4$
continuations per perception, for $N=ab=8$ completed trajectories per prompt.
The teacher returns top-$k$ token probabilities with $k=64$ on student-sampled
tokens. Teacher probabilities are used only in the perception-distillation
term; the separate token-level KL regularizer is disabled.

Policy optimization follows the DAPO-style decoupled clipping rule with
$\epsilon_{\mathrm{low}}=0.2$, $\epsilon_{\mathrm{high}}=0.28$, and dual-clip
$c=10$. The perception and reasoning loss coefficients are
$\lambda_{\mathrm{aw}}=0.1$ and $\lambda_{\mathrm{cot}}=1.0$, respectively.
Training runs for two epochs with an actor learning rate of $10^{-6}$, a
10-step warmup, and gradient-norm clipping at 1.0. A format bonus and
repetition/malformed-response penalties discourage late-training format
collapse. Table~\ref{tab:hparams} lists the complete configuration.

\subsection{Evaluation and Aggregation}

Unless a table states otherwise, evaluation uses \textsc{Avg@8}: eight
independent samples per question at temperature $1.0$ and top-$p=1.0$. For a
question $q$, the score is $8^{-1}\sum_{s=1}^{8}R_{qs}$; dataset accuracy is
the mean of these question-level scores. The reported macro average is the
unweighted mean over the eight datasets, so each benchmark contributes
equally regardless of dataset size. All values are percentages.

\subsection{Hardware and Runtime Measurement}

Training uses one node with eight NVIDIA B200 GPUs. Four GPUs serve student
training and rollout generation, and four serve teacher inference with tensor
parallel size four. Runtime is measured from steady-state end-to-end training
steps; checkpoint-saving steps are excluded. Because the 2B OPD run distills
longer joint responses whereas PCD distills the shorter separated perception
span, the measured time is an implementation-level comparison rather than a
claim of identical token counts.

\begin{table}[t]
    \centering
    \footnotesize
    \setlength{\tabcolsep}{4.5pt}
    \renewcommand{\arraystretch}{1.08}
    \resizebox{\columnwidth}{!}{%
    \begin{tabular}{ll}
        \toprule
        Setting & Value \\
        \midrule
        \multicolumn{2}{l}{\textit{Data / models}} \\
        Training dataset & Geo3K (\texttt{div} format) \\
        Model transfers & Qwen3-VL 8B$\to$2B; 32B$\to$8B \\
        Response spans & \texttt{<aware>} perception; \texttt{<cot>} reasoning \\
        \midrule
        \multicolumn{2}{l}{\textit{Optimization}} \\
        Train batch / mini-batch & $128$ / $32$ \\
        Actor learning rate & $1\times10^{-6}$ \\
        LR warmup / grad clip & $10$ steps / $1.0$ \\
        Clip ratio (low / high) & $0.2$ / $0.28$ (dual-clip $c{=}10$) \\
        Epochs / validation frequency & $2$ / every $5$ steps \\
        Maximum prompt / response length & $4096$ / $2048$ \\
        \midrule
        \multicolumn{2}{l}{\textit{Rollout}} \\
        Perceptions $a$ / continuations $b$ & $2$ / $4$ (budget $N{=}8$) \\
        Perception cap / stop sequence & $512$ / \texttt{</aware>} \\
        Sampling temperature / top-$p$ & $1.0$ / $1.0$ \\
        \midrule
        \multicolumn{2}{l}{\textit{PCD weighting}} \\
        Deficiency gate & $(1-\mathrm{PSR})\widetilde{\mathrm{KL}}$ \\
        KL normalization threshold & $0.3$ \\
        Base weight / boost $\alpha$ & $1.0$ / $1.0$ \\
        Mean-preserving normalization & enabled \\
        $\lambda_{\mathrm{aw}}$ / $\lambda_{\mathrm{cot}}$ & $0.1$ / $1.0$ \\
        Log-prob clamp & $-10.0$ \\
        \midrule
        \multicolumn{2}{l}{\textit{Teacher / regularization}} \\
        Teacher top-$k$ / tensor parallel & $64$ / $4$ \\
        Separate token-level KL loss & disabled \\
        Format bonus & $0.1$ \\
        Repetition / malformed penalty & $0.5$ / $0.5$ \\
        \bottomrule
    \end{tabular}%
    }
    \caption{Training and rollout configuration used unless stated otherwise.}
    \label{tab:hparams}
\end{table}

\section{Additional Analyses}
\label{sec:supp_analysis}

\subsection{Qualitative Perception-Correction Cases}

We compare standard on-policy distillation (OPD) against PCD on evaluation
questions from six representative benchmarks, selecting cases where all OPD
samples are wrong while PCD answers correctly.
Table~\ref{tab:cases_bench} shows one case for each represented benchmark. The pattern is
consistent across very different visual domains: OPD's answer fails because it
\emph{misreads the figure}---misjudging which angle a relation governs,
pairing the wrong vertices, or misreading the dominant visual content---whereas
PCD first commits a grounded \texttt{<aware>} perception and the reasoning
then follows correctly. Because these are eval-time benchmark prompts rather
than training prompts, the contrast reflects generalization of the corrected
perception, not memorization.

\begin{table*}[t]
    \centering
    \small
    \renewcommand{\arraystretch}{1.3}
    \begin{tabular}{@{}p{0.13\textwidth} p{0.23\textwidth} p{0.282\textwidth} p{0.282\textwidth}@{}}
        \toprule
        Dataset / diagram & Question & Standard OPD (wrong) & PCD (correct) \\
        \midrule
        \textbf{Geo3K}\newline
        \includegraphics[width=0.12\textwidth]{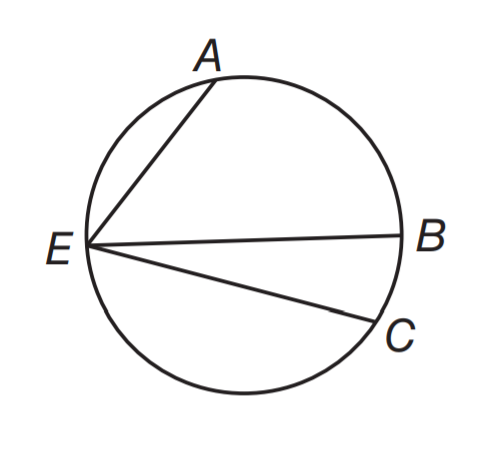} &
        $m\widehat{AC}=160^\circ$ and $m\angle BEC=38^\circ$;
        find $m\angle AEB$. &
        Finds arc $BC=76^\circ$ but mishandles the arc subtraction,
        committing $m\angle AEB=\boxed{62}$. &
        \emph{Perception:} circle with $A,B,C,E$; arc $AC{=}160^\circ$,
        $\angle BEC{=}38^\circ$. \emph{Reasoning:} arc $BC{=}2{\cdot}38{=}76^\circ$,
        arc $AB{=}160{-}76{=}84^\circ$, so $m\angle AEB{=}\tfrac{84}{2}=\boxed{42}$. \\
        \midrule
        \textbf{MathVista}\newline
        \includegraphics[width=0.10\textwidth]{Figures/cases_bench/MathVista.png} &
        $l_1\parallel l_2$; a $30^\circ$ set square rests with its right-angle
        vertex on $l_2$; given $\angle1=76^\circ$, find $\angle2$.
        (A)~$36^\circ$ (B)~$45^\circ$ (C)~$44^\circ$ (D)~$64^\circ$ &
        Misidentifies which angle the $30^\circ$ pairs with;
        chooses \textbf{(A)}~$36^\circ$. &
        \emph{Perception:} $l_1\parallel l_2$; a $30^\circ$--$90^\circ$ set square
        with its right-angle vertex on $l_2$, one leg parallel to $l_1$.
        \emph{Reasoning:} propagates $\angle1{=}76^\circ$ through the parallels with
        the $30^\circ$ to get \textbf{(C)}~$44^\circ$. \\
        \midrule
        \textbf{MathVerse}\newline
        \includegraphics[width=0.12\textwidth]{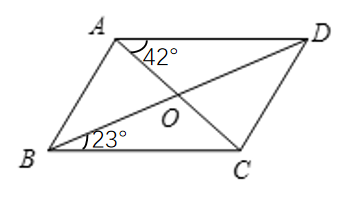} &
        Parallelogram $ABCD$ with $\angle A=42^\circ$, $\angle B=23^\circ$,
        diagonals meeting at $O$; find $\angle COD$.
        (A)~$61^\circ$ (B)~$63^\circ$ (C)~$65^\circ$ (D)~$67^\circ$ &
        Routes through the wrong triangle at $O$;
        chooses \textbf{(D)}~$67^\circ$. &
        \emph{Perception:} parallelogram, diagonals meet at $O$,
        $\angle A{=}42^\circ$, $\angle B{=}23^\circ$. \emph{Reasoning:} exterior-angle
        $\angle COD{=}\angle A{+}\angle B$, giving \textbf{(C)}~$65^\circ$. \\
        \midrule
        \textbf{MATH-Vision}\newline
        \includegraphics[width=0.10\textwidth]{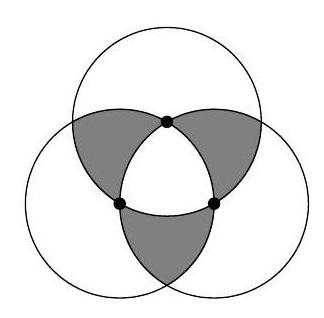} &
        Three radius-$2$ circles, each intersection point being the centre of
        the third circle; area of the grey zone?
        (A)~$\pi$ (B)~$3\pi$ (C)~$\tfrac{\pi}{2}$ (D)~$2\pi$ (E)~$4\pi$ &
        Misjudges the overlap geometry; chooses \textbf{(A)}~$\pi$. &
        \emph{Perception:} three $r{=}2$ circles, each intersection lying at another
        circle's centre (symmetric overlap). \emph{Reasoning:} the symmetric
        three-fold overlap area sums to \textbf{(D)}~$2\pi$. \\
        \midrule
        \textbf{We-Math}\newline
        \includegraphics[width=0.10\textwidth]{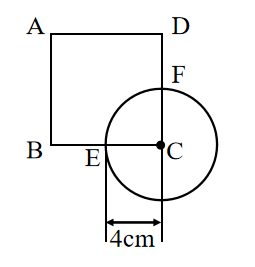} &
        A circle is centred at vertex $C$ of a square; find the central angle
        $\angle ECF$.
        (A)~$45$ (B)~$60$ (C)~$72$ (D)~$90$ (E)~none &
        Misplaces $E,F$ on the square; chooses \textbf{(C)}~$72^\circ$. &
        \emph{Perception:} circle centred at square-vertex $C$; $CE,CF$ are radii
        along two adjacent sides. \emph{Reasoning:} $\angle ECF$ equals the square's
        corner angle $=$ \textbf{(D)}~$90^\circ$. \\
        \midrule
        \textbf{MMStar}\newline
        \includegraphics[width=0.12\textwidth]{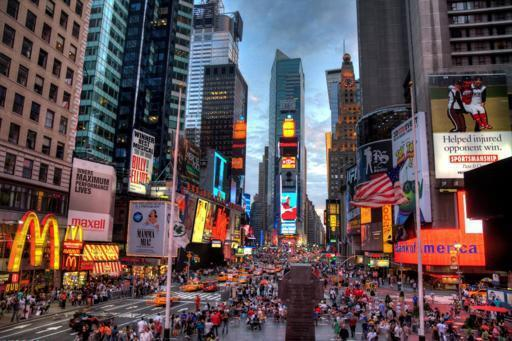} &
        What is the predominant feature in the image?
        (A)~Cars (B)~People (C)~Buildings (D)~Skyscrapers &
        Judges the crowd most numerous; chooses \textbf{(B)}~People. &
        \emph{Perception:} Times-Square scene; tall skyscrapers dominate the skyline
        (people and cars also present). \emph{Reasoning:} the skyscrapers are the most
        numerous, central feature $\Rightarrow$ \textbf{(D)}~Skyscrapers. \\
        \bottomrule
    \end{tabular}
    \caption{Paired OPD-versus-PCD cases across six representative benchmarks
    (evaluation-time
    \textsc{Avg@4} sampling, filtered to OPD all-wrong / PCD correct).
    Bracketed letters are the models' selected options; each PCD choice is the ground-truth
    answer. Across geometry, diagram reasoning, and natural-image
    understanding, OPD's error is a perception error that PCD's grounded
    \texttt{<aware>} step corrects.}
    \label{tab:cases_bench}
\end{table*}

In-domain, the same behavior appears on the Geo3K training rollouts.
Table~\ref{tab:cases} pairs OPD and PCD trajectories on identical training
prompts where every OPD trajectory is wrong: OPD pairs the wrong congruent
angles, mislabels corresponding sides, or mistakes a rectangle for a square,
while PCD's distilled perception makes the elementary algebra follow
correctly.

\begin{table*}[t]
    \centering
    \small
    \renewcommand{\arraystretch}{1.25}
    \begin{tabular}{@{}p{0.15\textwidth} p{0.19\textwidth} p{0.293\textwidth} p{0.293\textwidth}@{}}
        \toprule
        Diagram & Question & Standard OPD (wrong) & PCD (correct) \\
        \midrule
        \includegraphics[width=0.14\textwidth]{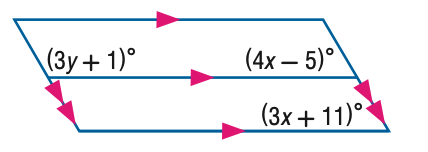} &
        Find $x$ (two parallel lines cut by transversals; angles $56^\circ$,
        $(3y{-}11)^\circ$, $4x^\circ$). &
        Misreads the angle relations and never resolves the linear pair;
        commits $x=\boxed{14}$. &
        \emph{Perception:} two parallel lines cut by transversals; angles
        $56^\circ$, $4x^\circ$, $(3y{-}11)^\circ$. \emph{Reasoning:} $4x^\circ$ is the
        linear pair of $56^\circ$, so $4x{+}56{=}180$ and $x=\boxed{31}$. \\
        \midrule
        \includegraphics[width=0.11\textwidth]{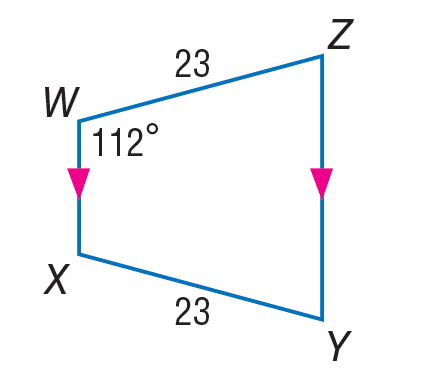} &
        Find $m\angle Z$ in isosceles trapezoid $WXYZ$
        ($m\angle W{=}112^\circ$). &
        Pairs the wrong congruent angles ($\angle W\!\cong\!\angle Z$),
        giving $m\angle Z=\boxed{112}$. &
        \emph{Perception:} isosceles trapezoid $WXYZ$, $WX{=}YZ{=}23$,
        $\angle W{=}112^\circ$. \emph{Reasoning:} $\angle W\!\cong\!\angle Y$, so the
        co-interior $\angle Z{=}180{-}112=\boxed{68}$. \\
        \midrule
        \includegraphics[width=0.10\textwidth]{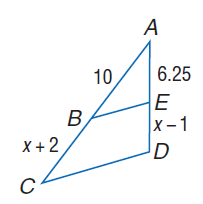} &
        Similar polygons; find $ED$ given $AB{=}10$, $AE{=}6.25$. &
        Mislabels which sides correspond, yielding $ED=\boxed{3.25}$. &
        \emph{Perception:} similar triangles with $AB{=}10$, $AE{=}6.25$,
        $AC{=}x{+}2$, $AD{=}x{-}1$. \emph{Reasoning:}
        $\tfrac{AB}{AE}{=}\tfrac{AC}{AD}$ from the correct correspondence gives
        $ED=\boxed{5}$. \\
        \midrule
        \includegraphics[width=0.14\textwidth]{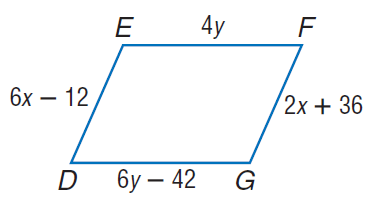} &
        Find $x$ so the quadrilateral is a parallelogram (diagonal segments
        $2x{+}3$, $5x$, $8y{-}36$, $4y$). &
        Pairs segments across different diagonals, giving $x=\boxed{-30}$. &
        \emph{Perception:} quadrilateral with diagonal segments $2x{+}3$, $5x$,
        $8y{-}36$, $4y$. \emph{Reasoning:} a parallelogram's diagonals bisect, so the
        two halves match: $2x{+}3{=}5x$ and $x=\boxed{1}$. \\
        \midrule
        \includegraphics[width=0.11\textwidth]{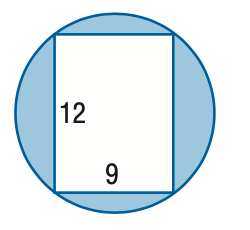} &
        Find the shaded area (a shape inscribed in a circle). &
        Perceives an inscribed \emph{square} of side $12$;
        area $=\boxed{82.2}$. &
        \emph{Perception:} a \emph{rectangle} $12\times9$ inscribed in the circle
        (not a square). \emph{Reasoning:} shaded $=$ circle $-$ rectangle
        $=\boxed{68.7}$. \\
        \bottomrule
    \end{tabular}
    \caption{Paired OPD-versus-PCD cases on identical Geo3K prompts
    (late-training rollouts, filtered to OPD all-wrong / PCD correct). Boxed
    values are the models' final answers; each PCD answer matches the ground
    truth. Across angle relations, similar figures, parallelograms, and area
    reasoning, OPD's error originates in perception and PCD's grounded
    \texttt{<aware>} step corrects it.}
    \label{tab:cases}
\end{table*}

\section{Rollout-Split and Gate-Rule Ablations}
\label{sec:supp_split_gate_ablation}

We further evaluate two design choices of the default PCD recipe with
controlled single-knob sweeps: the rollout-budget split $(a,b)$ under the
fixed trajectory budget $N=ab=8$, and the rule used to combine the two
deficiency witnesses. Each variant changes exactly one choice from the
default configuration ($2\times4$ rollout and multiplicative gate) while
retaining the same two-epoch training recipe, random seed, and
best-validation-checkpoint selection. We report in-domain Geo3K
\textsc{Avg@4} at temperature $T=1.0$, which directly measures the setting
used to construct the perception-level training signal.

Table~\ref{tab:supp_split_gate_ablation} shows that concentrating the budget
on either a single perception ($1\times8$) or a single continuation per
perception ($8\times1$) underperforms the default $2\times4$ allocation. The
multiplicative gate also outperforms the two single-witness rules and their
additive combination. Across held-out benchmarks, the \textsc{Avg@4} macro
averages of these variants remain within approximately 1.5 points on the
160-question, four-sample evaluation, making the in-domain result the more
discriminative comparison for this single-seed sweep.

\begin{table}[t]
    \centering
    \small
    \setlength{\tabcolsep}{6pt}
    \renewcommand{\arraystretch}{1.1}
    \begin{tabular}{lcc}
        \toprule
        Variant (one change from default) & Geo3K \textsc{Avg@4} & $\Delta$ \\
        \midrule
        \multicolumn{3}{l}{\textit{Rollout split $(a\times b)$, budget $N{=}8$}} \\
        $1\times8$ (single perception) & 40.3 & $-3.3$ \\
        \textbf{$2\times4$ (default)} & \textbf{43.6} & --- \\
        $4\times2$ & 39.7 & $-3.9$ \\
        $8\times1$ (single continuation) & 41.4 & $-2.2$ \\
        \midrule
        \multicolumn{3}{l}{\textit{Deficiency combination rule}} \\
        PSR-only & 38.3 & $-5.3$ \\
        KL-only & 41.1 & $-2.5$ \\
        Additive & 40.9 & $-2.7$ \\
        \textbf{Multiplicative (default)} & \textbf{43.6} & --- \\
        \bottomrule
    \end{tabular}
    \caption{Single-knob ablations of the default PCD recipe on in-domain
    Geo3K \textsc{Avg@4} ($T{=}1.0$). The balanced rollout allocation and
    two-witness multiplicative gate perform best within their respective
    sweeps.}
    \label{tab:supp_split_gate_ablation}
\end{table}

\section{Complete Transfer Results and Efficiency}
\label{sec:supp_full_results}

Table~\ref{tab:transfer_full} reports full-test-set
\textsc{Avg@8} ($n{=}8$, $T{=}1.0$) for the two Instruct-model transfer settings across
eight benchmarks. PCD uses the default configuration defined in
Section~\ref{sec:supp_protocol}; Standard OPD is the
plain on-policy-distillation baseline under the identical recipe. The
improvement is PCD minus Standard OPD on the same benchmark. PCD improves
the eight-benchmark macro average in both settings (${+}2.78$ for
$8\text{B}{\to}2\text{B}$, ${+}4.28$ for $32\text{B}{\to}8\text{B}$), with the
largest gain on the in-domain Geo3K benchmark; the two small regressions
(LogicVista, MMStar in the small-teacher setting) are reported honestly.
Table~\ref{tab:supp_runtime} separately reports the corresponding observed
end-to-end training time.

\begin{table*}[t]
    \centering
    \small
    \setlength{\tabcolsep}{6pt}
    \renewcommand{\arraystretch}{1.06}
    \resizebox{0.86\textwidth}{!}{%
    \begin{tabular}{lccc ccc}
        \toprule
        & \multicolumn{3}{c}{8B teacher $\rightarrow$ 2B student}
        & \multicolumn{3}{c}{32B teacher $\rightarrow$ 8B student} \\
        \cmidrule(lr){2-4}\cmidrule(lr){5-7}
        Benchmark & OPD & PCD & $\Delta$ & OPD & PCD & $\Delta$ \\
        \midrule
        Geo3K$^\dagger$ & 30.82 & \textbf{42.12} & \textbf{+11.30}
            & 42.51 & \textbf{59.65} & \textbf{+17.14} \\
        MMMU-Pro & 45.78 & \textbf{47.59} & +1.81
            & 59.05 & \textbf{61.45} & +2.40 \\
        We-Math & 59.78 & \textbf{62.44} & +2.66
            & 72.19 & \textbf{73.92} & +1.73 \\
        MathVista & 61.55 & \textbf{62.45} & +0.90
            & 74.50 & \textbf{75.79} & +1.29 \\
        MathVerse & 33.14 & \textbf{37.72} & +4.58
            & 44.97 & \textbf{50.10} & +5.13 \\
        MATH-Vision & 20.19 & \textbf{24.01} & +3.82
            & 34.33 & \textbf{39.88} & +5.55 \\
        MMStar & \textbf{59.92} & 59.06 & $-0.86$
            & 69.44 & \textbf{70.43} & +0.99 \\
        LogicVista & \textbf{44.84} & 42.89 & $-1.95$
            & 58.51 & \textbf{58.57} & +0.06 \\
        \midrule
        \textbf{Macro avg} & 44.50 & \textbf{47.28} & \textbf{+2.78}
            & 56.94 & \textbf{61.22} & \textbf{+4.28} \\
        \bottomrule
    \end{tabular}%
    }

    \caption{Complete full-set \textsc{Avg@8} results ($n{=}8$, $T{=}1.0$,
    $0$--$100$ scale). $^\dagger$Geo3K is the in-domain training benchmark.}
    \label{tab:transfer_full}
\end{table*}

\begin{table}[t]
    \centering
    \small
    \resizebox{\columnwidth}{!}{%
    \begin{tabular}{llcc}
        \toprule
        Student & Teacher & OPD (s/step) & PCD (s/step) \\
        \midrule
        2B Instruct & 8B Instruct & 183.9 & 132.5 \\
        2B Instruct & 8B Instruct-GRPO & -- & 356 \\
        8B Instruct & 32B Instruct & 417 & 412 \\
        \bottomrule
    \end{tabular}%
    }
    \caption{Observed end-to-end step time on 8$\times$B200. Runtime averages
    exclude checkpoint-saving steps. The 2B timing difference also reflects
    the shorter separated perception span distilled by the PCD implementation.
    The Instruct-GRPO teacher configuration has no matched standard-OPD run.}
    \label{tab:supp_runtime}
\end{table}

\section{Optional All-Wrong Exploration}
\label{sec:supp_optional_exploration}

We also implemented an exploratory response to prompts for which every sampled
trajectory is incorrect. It is \emph{disabled in all reported results} and is
not part of evaluated PCD. The allocation can move from $2\times4$ to
$4\times2$ and then to $8\times1$, trading PSR precision for perception
diversity. Each allocation contains eight trajectories, but sequential retries
consume an additional eight trajectories each; only selecting one split before
generation preserves fixed per-prompt compute.

At the final $8\times1$ allocation, an optional exploratory perception weight
is
\begin{equation}
    w_i^{\mathrm{explore}}
    =-w_{\max}^{\mathrm{explore}}
      \operatorname{clip}\!\left(
      1-\frac{\mathrm{KL}_i}{\tau_{\mathrm{KL}}},0,1\right),
\end{equation}
with $w_{\max}^{\mathrm{explore}}=0.1$ and
$\tau_{\mathrm{KL}}=5.0$. The small negative weight moves away from a
teacher-aligned perception mode that nevertheless failed. The operation is
gated by \texttt{dyn\_rollout\_enable} and
\texttt{explore\_aware\_enable}; because it changes the sign of distillation
and has not been isolated, it remains a hypothesis for future study.

\section{Extended Discussion and Limitations}

\paragraph{What PCD changes.}
PCD defines the unit and evidence on which a distillation weight is computed:
it treats perception as a sampled latent action, estimates downstream value
from shared continuations, and combines that value with teacher disagreement.
Mean preservation separates supervision allocation from total imitation.

\paragraph{When the signal is informative.}
PCD is most informative when visual evidence is pivotal, reasoning is
stochastic, and the teacher is perceptually stronger. It is weaker on
text-dominated questions or shared teacher--student failures. Disagreement is
therefore a witness of teacher-correctable deficiency, not a correctness label.

\paragraph{Scope of the evidence.}
The endpoint comparisons and single-seed sweeps support separated sampling,
the balanced rollout allocation, and multiplicative adaptive weighting, but
the PSR-only, KL-only, additive, and multiplicative gates have not yet been
compared under matched multi-seed training. Multi-seed sensitivity analysis,
additional model families, and broader training domains remain necessary for
stronger generality claims.


\end{document}